\documentclass[10pt,twocolumn,letterpaper]{article}

\usepackage[pagenumbers]{cvpr} 

\usepackage{graphicx}
\usepackage{amsmath}
\usepackage{amssymb}
\usepackage{booktabs}

\usepackage{multirow}
\usepackage{tabularx}
\usepackage{etoolbox,siunitx}
\robustify\bfseries
\usepackage{enumitem}
\usepackage{float}
\setlist[itemize]{noitemsep, topsep=0pt}

\usepackage{comment}

\usepackage[table]{xcolor}

\usepackage{adjustbox}

\usepackage[super]{nth}

\DeclareMathOperator*{\argmin}{\arg\!\min}

\newif\ifdrafting
\draftingtrue 
\ifdrafting
    \newcommand{\ds}[1]{{\leavevmode\color[rgb]{1,0,0}[Deqing: #1]}}
    \newcommand{\as}[1]{{\leavevmode\color[rgb]{.2,1.0,.2}[Austin: #1]}}
    \newcommand{\cih}[1]{{\leavevmode\color[rgb]{0,0.5,0}[Charles: #1]}}
    \newcommand{\jh}[1]{{\leavevmode\color[rgb]{0.87, 0.0, 1.0}[Junhwa: #1]}}
    \newcommand{\todo}[1]{{\leavevmode\color[rgb]{1,0,0}#1}}
    \newcommand{\cindy}[1]{{\leavevmode\color[rgb]{0,0.8,0.8}[Cindy: #1]}}

\else
    \newcommand{\ds}[1]{}
    \newcommand{\as}[1]{}
    \newcommand{\cih}[1]{}
    \newcommand{\jh}[1]{}
    \newcommand{\todo}[1]{}
    \newcommand{\cindy}[1]{}
\fi

\newcommand{\ignore}[1]{}
\makeatletter
\newcommand{\printfnsymbol}[1]{%
        \textsuperscript{\@fnsymbol{#1}}%
}
\makeatother

\newcommand{\myparagraph}[1]{\smallskip\noindent\textbf{#1}}

\newcommand{\autoflow}{AF}
\newcommand{\selfautoflow}{S-AF}

\graphicspath{{figure}, {images}, {example}}

\usepackage[pagebackref,breaklinks,colorlinks,bookmarks=false,citecolor=blue,linkcolor=blue]{hyperref}

\usepackage[capitalize]{cleveref}

\crefname{section}{Sec.}{Secs.}
\Crefname{section}{Section}{Sections}
\Crefname{table}{Table}{Tables}
\crefname{table}{Tab.}{Tabs.}
\usepackage{color, colortbl}
\definecolor{LightGray}{gray}{0.95}
\definecolor{LightCyan}{rgb}{0.88,1,1}

\begin{document}

\title{
Self-supervised AutoFlow
}

\author{Hsin-Ping Huang$^{1,2}$, Charles Herrmann$^1$, Junhwa Hur$^1$, Erika Lu$^1$, \\
Kyle Sargent$^1$, Austin Stone$^1$, Ming-Hsuan Yang$^{1,2}$, Deqing Sun$^1$\\
$^1$Google Research~~~$^2$University of California, Merced
}

\twocolumn[{
\renewcommand\twocolumn[1][]{#1}
\maketitle
\begin{center}
    \captionsetup{type=figure}
    \includegraphics[width=\textwidth]{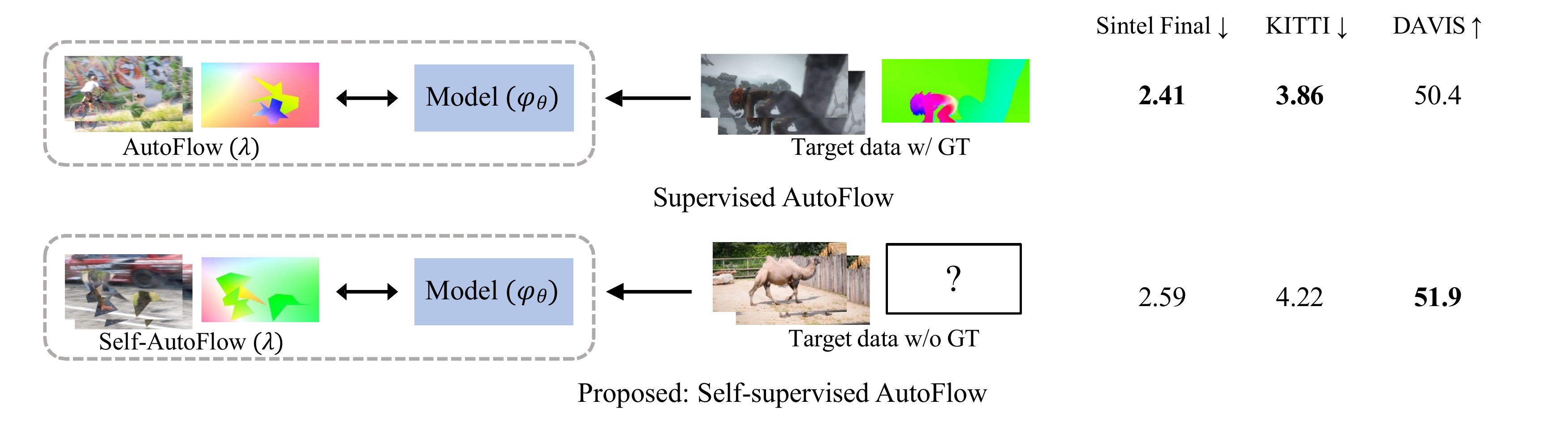}
    \captionof{figure}
    {{\bf Self-supervised AutoFlow} learns to generate an optical flow training set through self-supervision on the target domain. It performs comparable to supervised AutoFlow on Sintel and KITTI without requiring ground truth (GT) and learns a better dataset for real-world DAVIS, where GT is not available. We report optical flow accuracy on Sintel and KITTI, and keypoint propagation accuracy on DAVIS.
}
\end{center}
}]

\begin{abstract}

Recently, AutoFlow has shown promising results on learning a training set for optical flow, but requires ground truth labels in the target domain to compute its search metric. Observing a strong correlation between the ground truth search metric and self-supervised losses, we introduce self-supervised AutoFlow to handle real-world videos without ground truth labels. Using self-supervised loss as the search metric, our self-supervised AutoFlow performs on par with AutoFlow on Sintel and KITTI where ground truth is available, and performs better on the real-world DAVIS dataset. We further explore using self-supervised AutoFlow in the (semi-)supervised setting and obtain competitive results against the state of the art.

\end{abstract}

\section{Introduction}
\label{sec:intro}

\begin{flushright} 
\emph{Data is the new oil.} ---  Clive Humby, 2006~\cite{humby2006data} 
\end{flushright}

This well-known analogy not only foretold the critical role of data for developing AI algorithms in the last decade but also revealed the importance of \textit{data curation}. Like refined oil, data must be carefully curated to be useful for AI algorithms to succeed. For example, one key ingredient for the success of AlexNet~\cite{Alexnet} is ImageNet~\cite{russakovsky2015imagenet}, a large dataset created by extensive manual labeling. 

The manual labeling process, however, is either not applicable or difficult to scale to many low-level vision tasks, such as optical flow. A common practice for optical flow is to pre-train models using large-scale synthetic datasets, \eg, FlyingChairs \cite{FlowNet} and FlyingThings3D \cite{Mayer2018}, and then fine-tune them on limited in-domain datasets, \eg, Sintel~\cite{ButlerECCV2012} or KITTI~\cite{KITTI2015}. While this two-step process works better than directly training on the limited target datasets, there exists a domain gap between synthetic data and the target domain. 

To narrow the domain gap, AutoFlow~\cite{sun2021autoflow} learns to render a training dataset to optimize performance on a target dataset, obtaining superior results on Sintel and KITTI where the ground truth is available. 
As obtaining ground truth optical flow for most real-world data is still an open challenge, it is of great interest to remove this dependency on ground truth to apply AutoFlow to real-world videos.

In this paper, we introduce a way to remove this reliance by connecting learning to render with another independent line of research on optical flow, self-supervised learning (SSL). SSL methods for optical flow~\cite{Zhong2019UnsupervisedDE,DDFlow,SelFlow,im2020unsupervised,liu2020learning} use a set of self-supervised losses to train models using only image pairs in the target domain. 
We observe a strong correlation between these self-supervised losses and the ground truth errors, as shown in~\cref{fig:correlation}. This motivates us to connect these two lines of research by adopting self-supervised losses as a search metric for AutoFlow\cite{sun2021autoflow}, calling our approach ``Self-supervised AutoFlow''. 

Self-supervised AutoFlow obtains similar performance to AutoFlow on Sintel~\cite{ButlerECCV2012} and KITTI~\cite{KITTI2015}, and it can learn a better dataset for the real-world DAVIS data \cite{DAVIS} where ground truth is not available. To further narrow the domain gap between synthetic data and the target domain, we also explore new ways to better synergize techniques from learning to render and self-supervised learning.

Numerous self-supervised methods still rely on pre-training on a synthetic dataset. Our method replaces this pre-training with supervised training on self-supervised AutoFlow data generated using self-supervised metrics. This new pipeline is still self-supervised and obtains competitive performance among all self-supervised methods. We further demonstrate that our method provides a strong initialization for supervised fine-tuning and obtains competitive results against the state of the art. 

We make the following main contributions: 
\begin{itemize}
    \item We introduce self-supervised AutoFlow to learn to render a training set for optical flow using self-supervision on the target domain, connecting two independently studied directions for optical flow: learning to render and self-supervised learning. 
    \item Self-supervised AutoFlow performs competitively against AutoFlow~\cite{sun2021autoflow} that uses ground truth on Sintel and KITTI and better on DAVIS where ground truth is not available. 
    \item We further analyze self-supervised AutoFlow in semi-supervised and supervised settings and obtain competitive performance against the state of the art.
\end{itemize}

\section{Related Work}
\label{sec:related}

\myparagraph{CNN architectures for optical flow.}
Recent advances in deep learning and synthetic datasets have contributed to the development of numerous optical flow architectures.
Early work introduces basic designs using U-Net \cite{ronneberger2015u,FlowNet,Flownet2} or an image pyramid \cite{spynet2017}.
PWC-Net \cite{Sun2018PWCNet}, concurrently with LiteFlowNet \cite{Hui_2018_CVPR}, introduces an advanced design based on well-established domain knowledge (\eg,~pyramid, warping, and cost volume).
RAFT \cite{RAFT} further advances architecture designs based on a full 4D cost volume with a recurrent optimizer, which significantly improves the accuracy and encourages many follow-up methods \cite{jiang2021learning,zhang2021separable,xiao2020learnable,jahedi2022multi,sun2022disentangling}, followed by recent attention-based designs \cite{xu2021flow1d,huang2022flowformer,sui2022craft,jaegle2021perceiver} as well. As our main focus is on the dataset, we adopt the widely-used RAFT architecture in our experiments. 

\myparagraph{Self-supervised optical flow.}
Supervised approaches may not generalize well to real-world domains where annotations are difficult to obtain.
To overcome the limitation, self-supervised approaches~\cite{ahmadi2016unsupervised,jason2016back2basics,ren2017unsupervised,Zhong2019UnsupervisedDE} directly train the networks on the target data with hand-crafted self-supervised losses \cite{meister2018unflow,im2020unsupervised,wang2018occlusion,DDFlow,SelFlow,liu2020learning}.
UFlow~\cite{jonschkowski2020matters} systematically analyzes the effect of various loss designs on the accuracy and proposes an optimized combination for the best accuracy.
SMURF~\cite{stone2021smurf} presents a self-supervised method based on the RAFT~\cite{RAFT} architecture and proposes several technical designs such as the sequence loss, full image warping, heavy augmentation, and multi-frame training.
In this paper, we find that there is a strong correlation between self-supervised loss and ground truth errors, which inspires us to employ the self-supervised loss as a search metric for synthetic dataset learning. We further explore ways to synergize self-supervised methods and learning to render for better performance in the self-supervised setting. 

\myparagraph{Semi-supervised optical flow.}
To benefit from training on both labeled (out-of-domain) data and target domains, semi-supervised approaches propose to reduce a domain gap between datasets by using a GAN \cite{goodfellow2014generative, lai2017semi}, to learn a conditional prior from labeled data \cite{yang2018conditional}, to benefit from a small fraction of labels by active learning \cite{yuan2022optical}, or to adapt to the target domain through knowledge distillation \cite{im2022semi}.
SemiFlow \cite{han2022realflow} introduces an iterative approach that generates a training dataset in the real-world domain using a pre-trained model and trains the model using the generated dataset.
These methods usually rely on models trained on datasets designed manually, \eg, FlyingChairs and FlyingThings3D. Our work shows that using the self-supervised AutoFlow dataset can further improve performance and, more importantly, remove manual design processes from the entire pipeline. 

\myparagraph{Training datasets for optical flow.}
Due to the difficulty of constructing large-scale real-world annotated datasets for optical flow, synthetic data (\eg~FlyingChairs~\cite{FlowNet}, FlyingThings3D~\cite{Mayer2018}, Kubric~\cite{greff2022kubric}) have been widely used as standard (pre-)training datasets.
However, these datasets are generated without consideration of a target domain, so the domain gap always exists between the training and target domain, \eg, MPI Sintel~\cite{ButlerECCV2012} or VIPER~\cite{richter2017playing} \vs KITTI~\cite{KITTI2015}. 

Two works have introduced a training dataset generation pipeline based on real-world images. 
Depthstillation~\cite{aleotti2021learning} synthesizes an image at an arbitrarily rotated view from a still image and provides optical flow ground truth between the images. 
RealFlow~\cite{han2022realflow} synthesizes an intermediate frame between two frames given an estimated flow. The synthesis is controlled to have motion statistics similar to the target dataset. 
Both methods require off-the-shelf monocular depth methods \cite{ranftl2021vision,ranftl2020towards} and a hole-filling method to minimize artifacts on synthesized images. Furthermore, there is no guarantee that models trained on the synthesized datasets will perform optimally on the target domain. 

AutoFlow~\cite{sun2021autoflow} proposes a learning-to-render pipeline that learns dataset-rendering hyperparameters to optimize the optical flow accuracy on the target domain. 
Our method follows a similar direction, but unlike AutoFlow~\cite{sun2021autoflow}, does not require ground truth labels on the target domain. Instead, it uses a self-supervised search metric to update the rendering hyperparameters, making it applicable to any target domain without available ground truth.

\section{Approach}
\label{sec:approach}

Given an {unlabeled} target dataset $\mathbf{D}_\text{target}$, we aim to learn a synthetic dataset $\mathbf{D}_\text{auto}$ that approximately optimizes the performance in the target domain. To this end, we introduce self-supervised AutoFlow, which connects two independent research directions: {(i)} learning to render training datasets and {(ii)} self-supervised learning of optical flow (\cref{sec:autoflow}).
Then, given the generated dataset $\mathbf{D}_\text{auto}$ with ground truth and the unlabeled target dataset $\mathbf{D}_\text{target}$, our method trains an optical flow network $\phi_\theta$ using self-supervision to further adapt to the target domain (\cref{sec:combine}). 
The whole pipeline is fully self-supervised and does not require any ground truth optical flow from the target domain. 

\subsection{Preliminary: (Supervised) AutoFlow}

AutoFlow~\cite{sun2021autoflow} uses a layered approach to render a training dataset.
The rendering pipeline uses a set of hyperparameters $\lambda$ that control visual properties of foreground objects and the background (\eg the number of moving objects, object shape, size, motion, \etc) and their visual effects (\eg motion blur, fog, \etc) that appear in the rendered dataset. 
In a pre-defined hyperparameter search space $\Lambda$, an optimization process searches for an optimal set of hyperparameters $\lambda^*$ such that $\phi_\theta{(\lambda)}$, an optical flow network trained on a rendered dataset with the parameters $\lambda$, minimizes a pre-defined search metric $\Omega$ on the target dataset:
\begin{equation}
    \lambda^* = \argmin_{\lambda \in \Lambda} {\Omega} \left (\phi_\theta{(\lambda)} \right ).
    \label{eq:param_opt}
\end{equation}

AutoFlow~\cite{sun2021autoflow} uses average end-point error (AEPE) for the search metric ${\Omega}$ that measures the accuracy between available ground truth in the target dataset and estimated optical flow from the trained model $\phi_\theta{(\lambda)}$. Despite promising results on Sintel and KITTI, AutoFlow cannot be applied to real-world data that do not have optical flow annotations.

\begin{figure}[t]
    \centering
    \includegraphics[width=0.9\columnwidth]{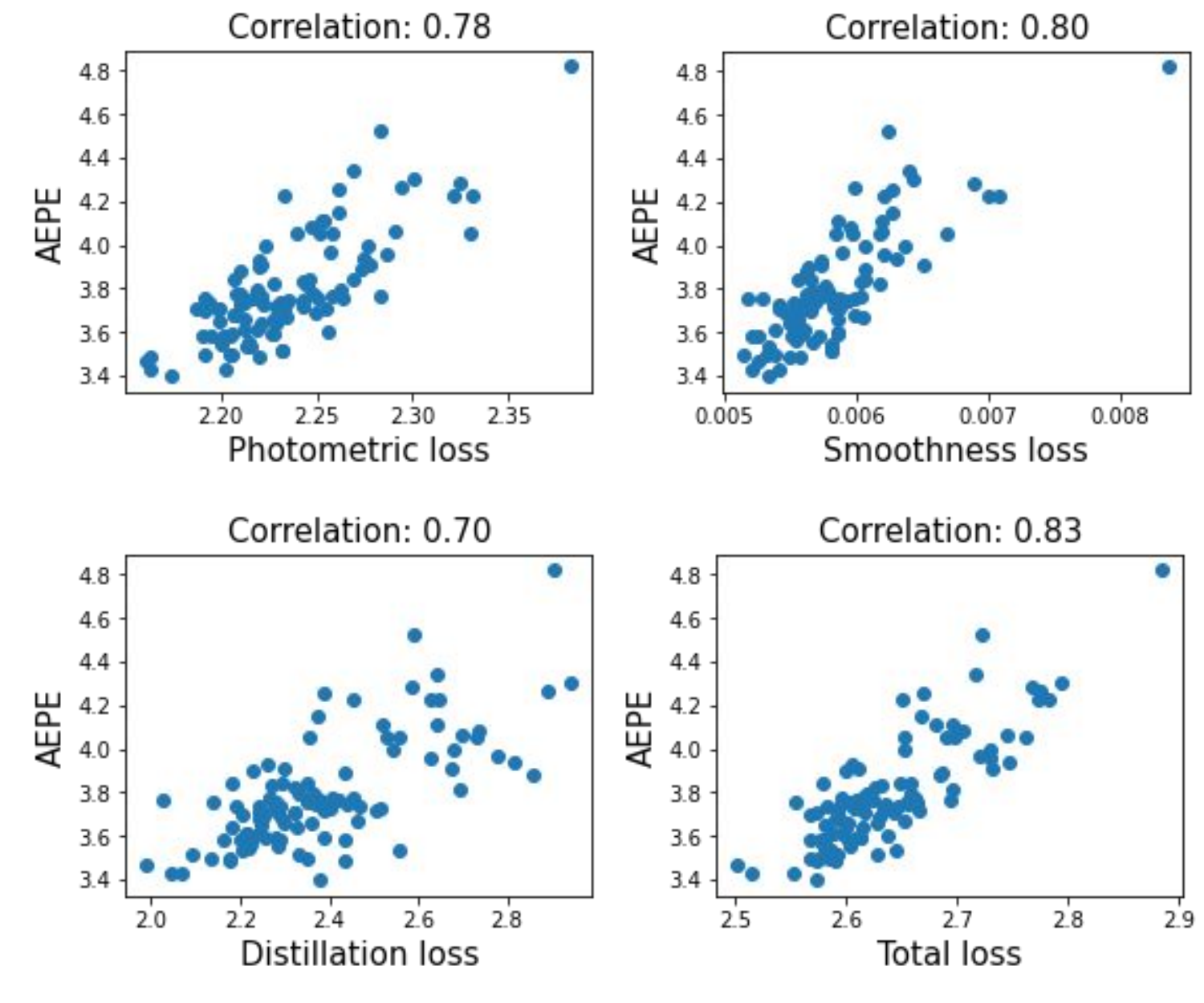}
    \caption{\textbf{Strong correlation} between ground truth error metric (AEPE) and self-supervised losses. We evaluate a set of RAFT models trained on supervised AutoFlow~\cite{sun2021autoflow} datasets using the ground truth average end-point error (AEPE) and self-supervised losses averaged on the Sintel Final data. Each point in the plots corresponds to the performance of one model.
    }
    \label{fig:correlation}
    \vspace{-0.5em}
\end{figure}

\subsection{Self-supervised AutoFlow}
\label{sec:autoflow}
\myparagraph{Motivation.}
To remove AutoFlow's dependence on in-domain ground truth, we look for inspiration from another line of research: self-supervised learning for optical flow. 
In particular, the recent SMURF \cite{stone2021smurf} outperforms the supervised PWC-Net \cite{Sun2018PWCNet} (the state of the art 4 years ago) on Sintel and KITTI, suggesting that its self-supervised loss is highly correlated with the ground truth errors and could be a good proxy metric for learning optical flow. 

To this end, we analyze the correlation between the ground truth average end-point error (AEPE) metric and SMURF's~\cite{stone2021smurf} self-supervised loss on Sintel using the trained models during the hyperparameter search of the supervised AutoFlow \cite{sun2021autoflow}, shown in~\cref{fig:correlation}.
Each point in the plot corresponds to a RAFT model trained on a supervised AutoFlow dataset, with its AEPE on the Sintel Final split ($y$ axis) and the self-supervised loss ($x$ axis) that consists of a photometric loss, smoothness loss, and distillation loss.
As shown in the plots, lower self-supervised losses correspond to lower AEPEs, and the correlation between the two signals increases when multiple losses are combined (\ie total loss).
This observation suggests that the self-supervised loss can also serve as a reliable proxy search metric and motivates our Self-supervised AutoFlow.

\myparagraph{Self-supervised search metric.} 
Our work extends the applicability of AutoFlow and presents Self-supervised AutoFlow (Self-AutoFlow or S-AF) which enables rendering a training dataset for a target domain by relying on the self-supervision loss metrics.
We define our search metric $\Omega$ using a self-supervised loss which consists of three terms, a photometric loss $\mathcal{L}_\text{photo}$, a smoothness loss $\mathcal{L}_\text{smooth}$, and a distillation loss $\mathcal{L}_\text{distill}$,
\begin{equation}
\Omega_\text{\selfautoflow} (\phi_\theta{(\lambda)}) = \mathcal{L}_\text{photo} + \omega_{\text{smooth}} \mathcal{L}_\text{smooth} + \omega_{\text{distill}} \mathcal{L}_\text{distill},
\label{eq:search_metric_ours}
\end{equation}

\noindent 
where each loss function follows that of SMURF's \cite{stone2021smurf} and $\omega_*$ are weighting coefficients. 
The input to each loss term is a pair of input images and an estimated optical flow from a trained model $\phi_\theta{(\lambda)}$, and are omitted for brevity.

The photometric loss $\mathcal{L}_\text{photo}$ penalizes the difference of corresponding pixels between input images $\textbf{I}_t$ and $\textbf{I}_{t+1}$. $\textbf{I}_{t+1}$ is differentiably warped into $\textbf{I}_t$  using the predicted optical flow, $\textbf{W}_t$, and following \cite{zabih1994non}, a Hamming distance of ternary-census-transformed image patches of corresponding pixels is used to compute the photometric loss with respect to $\textbf{W}_t$.
The smoothness loss $\mathcal{L}_\text{smooth}$ uses the $k^{\textnormal{th}}$ order edge-aware smoothness to encourage continuity of the predicted optical flow field while allowing for discontinuity on edges. 
The distillation loss $\mathcal{L}_\text{distill}$ (\ie,~`self-supervision loss' in SMURF \cite{stone2021smurf}) applies a loss between a prediction on original images from a teacher model and a prediction on augmented and cropped images from a student model. 
As there is no backpropagation to the model in the search of AutoFlow, the search metric uses only the final, instead of all intermediate, flow prediction of RAFT.

\myparagraph{Mixed datasets.}
Despite the high correlation between self-supervised loss and the ground truth error metric, there is no guarantee that the top candidate returned by self-supervised AutoFlow is the optimal set of hyperparameters according to the ground truth. 
To increase robustness, we choose the top-3 hyperparameter sets returned by self-supervised AutoFlow, generate a set of images with ground truth from each hyperparameter set, and mix them equally to form our final self-supervised AutoFlow dataset $\mathbf{D}_\text{auto}$.
Empirically, we find that mixing the datasets decreases the likelihood of sampling a poor-performing AutoFlow hyperparameter and generally improves the robustness of the algorithm.

\myparagraph{Discussion.}
There is a significant difference between learning a training set using self-supervised search metrics and self-supervised learning for optical flow.
Self-supervised learning of optical flow involves training directly on a target dataset using self-supervised proxy losses. 
Gradients from the losses are directly backpropagated to update the model parameters.
In contrast, our self-supervised AutoFlow approach optimizes hyperparameters for rendering a training dataset and trains the model on the dataset generated by the hyperparameters.
The high correlation between the self-supervised loss and the ground truth error makes the Self-AutoFlow dataset almost as good as the AutoFlow dataset.
The rendering pipeline can serve as an inductive bias for self-supervised learning and provide ground truth for complex scenes, such as occlusions and motion blur, that models trained on self-supervised losses tend to fail.

\subsection{Combining Self-supervised AutoFlow with Self-supervised Optical Flow}
\label{sec:combine}
Given the AutoFlow dataset $\mathbf{D}_\text{auto}$ learned from the self-supervised search metric, we further combine two data sources for training: (i) the self-supervised AutoFlow data $\mathbf{D}_\text{auto}$ and (ii) a target dataset without ground truth $\mathbf{D}_\text{target}$.
Specifically, we first pre-train the model on $\mathbf{D}_\text{auto}$ and then self-supervised fine-tune the model on the target dataset $\mathbf{D}_\text{target}$, based on a training protocol from SMURF~\cite{stone2021smurf}. 

\myparagraph{Self-supervised fine-tuning.}
This stage is to further adapt the model to the unlabeled target domain (\ie raw videos).
We use the same self-supervised loss from \cref{eq:search_metric_ours}. 
\begin{equation}
\mathcal{L} = \mathcal{L}_\text{photo} + \omega_{\text{smooth}} \mathcal{L}_\text{smooth} + \omega_{\text{distill}} \mathcal{L}_\text{distill}.
\label{eq:ft_self_sup_loss}
\end{equation}

\myparagraph{Multi-frame fine-tuning.}
After fine-tuning on the target domain with the self-supervised loss in \cref{eq:ft_self_sup_loss}, we further apply the multi-frame fine-tuning from SMURF \cite{stone2021smurf}.
Given a triplet of input frames ($\mathbf{I}_{t-1}$, $\mathbf{I}_{t}$, and $\mathbf{I}_{t+1}$), SMURF predicts bi-directional flow ($(\mathbf{I}_t \rightarrow \mathbf{I}_{t-1})$ and $(\mathbf{I}_t \rightarrow \mathbf{I}_{t+1})$) and generates pseudo ground truth for the forward flow $\textbf{W}_\text{pseudo}$ that includes more reliable estimation on occluded pixels through occlusion detection and inpainting using a shallow CNN. 
Then, we apply the following sequence loss from RAFT \cite{RAFT}, which applies the $l_1$ loss ($\rho_F$) on each
$n^{\textnormal{th}}$ intermediate output $\mathbf{W}^n$ with a decay factor $\gamma$,
\begin{equation}
    \begin{split}
    \mathcal{L} &= 
    \sum_{n}{\gamma^{N-n}\rho_F(\mathbf{W}_\text{pseudo}-\mathbf{W}^n)}.
    \end{split}
\label{eq:ft_multiframe}
\end{equation}

\begin{figure*}[ht]
    \centering
    \includegraphics[width=1.0\textwidth]{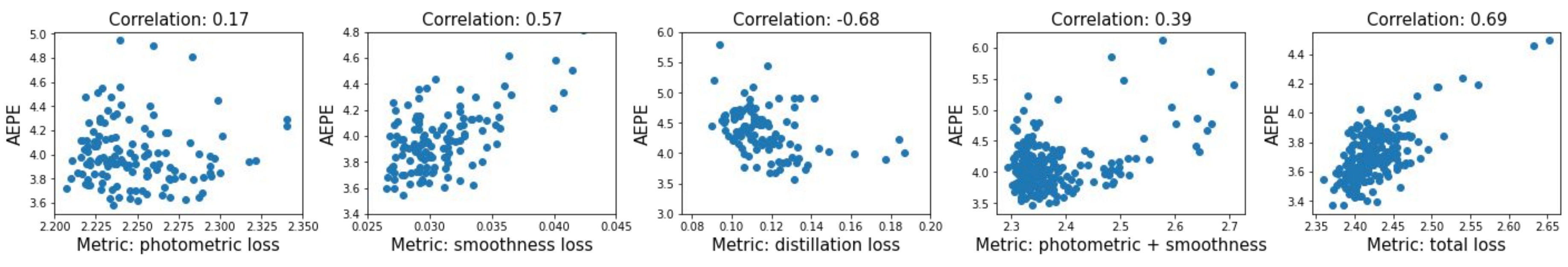}
    \caption{\textbf{Ablation study of self-supervised search metric.} None of the individual terms of the standard self-supervised loss, when used as a search metric, is strongly correlated with AEPE on the target dataset. Only the combination of all three terms leads to a strong correlation between the search metric and the AEPE. Each point here denotes an AEPE of a model trained on a generated dataset searched by a self-supervised search metric, whereas \cref{fig:correlation} shows the supervised AutoFlow models that use AEPE for the dataset parameter search.
    }
    \label{fig:ablation}
    \vspace{-0.5em}
    
\end{figure*}

\section{Experiments}
\label{sec:results}

\subsection{Experimental setup}
We use RAFT~\cite{RAFT} as the backbone architecture.
For the self-supervised hyperparameter search, we train 16 models in parallel using 96 NVIDIA P100 GPUs.
At each search iteration, we train models for a short amount of steps, evaluate them on our search metric (\cref{eq:search_metric_ours}), and update the hyperparameters. 
We conduct 8 search iterations, which results in 16$\times$8 total models for the search.
We use the Adam optimizer ($\beta_1{=}0.9$, $\beta_2{=}0.999$) with a learning rate of 0.0001 and a one-cycle learning rate schedule \cite{smith2019super}.
Our method has the same theoretical complexity as AutoFlow. However, in practice, we reduce the computation cost by nearly 60\% by using fewer training steps (80k).

For each target domain, we conduct a separate search and render a separate dataset. 
After pre-training on the rendered dataset, we further fine-tune the model with the self-supervised loss (\cref{eq:ft_self_sup_loss}) on each target dataset, followed by multi-frame fine-tuning (\cref{eq:ft_multiframe}).
We use a learning rate of 0.0002 with an exponential decay during the last 20\% of steps.
At inference time, we use the established evaluation scheme for each domain. \cref{tab:pretrain} follows AutoFlow's, which uses a fixed resolution during inference (Sintel: 448$\times$1024, KITTI: 640$\times$640). \cref{tab:sota} and \cref{tab:general} follow SMURF's, which uses resolutions that perform the best
on the training set (Sintel: 384$\times$1024, KITTI: 424$\times$952).

\begin{table}[t]
\centering
\caption{ \textbf{Comparison of (self-)supervised pre-training approaches.} Our Self-AutoFlow (\selfautoflow) outperforms FlyingChairs pre-training and is competitive with supervised AutoFlow (AF) which is learned from ground truth annotations. {\bf Bold} indicates the best number. ``\autoflow~X'', ``\autoflow-mix X'' or ``\selfautoflow~X'' indicates that \autoflow~or \selfautoflow~is learned for the dataset X. Numbers in parentheses indicate the number of training steps.
}
\scriptsize
\begin{tabularx}{0.9\columnwidth}{@{} X *{3}{ @{\hskip 1.5em} c} @{}}

\multirow{2}{*}{Dataset and Method} & Sintel Clean  & Sintel Final  & KITTI \\ 
& (AEPE~$\downarrow$) & (AEPE~$\downarrow$) & (AEPE~$\downarrow$) \\
\midrule

\textbf{Supervised} & & & \\
RAFT Chairs \cite{RAFT} & 2.27 & 3.76 & 7.63\\ 
\autoflow~Sintel (3.2M)~\cite{sun2022disentangling} & \bf 1.74 & \bf 2.41 & 4.18\\
\autoflow-mix Sintel (3.2M) & 1.85 & 2.53 & 3.92 \\
{\autoflow~KITTI (0.8M)~\cite{sun2021autoflow}} & 2.09 & 2.82 & 4.33 \\
\autoflow-mix KITTI (0.8M)  & 1.87 & 2.77 & \bf 3.86 \\
 
 \midrule

\textbf{Self-supervised} & & &\\
SMURF Chairs~\cite{stone2021smurf} & 2.19 & 3.35 & 7.94 \\
\selfautoflow~Sintel (3.2M)  &  \textbf{1.83} &	\textbf{2.59} & 5.22 \\
\selfautoflow~KITTI (0.2M) & 2.20 &  3.01 & 4.58 \\
\selfautoflow~KITTI (0.8M) &  1.99 & 3.00 & 4.29 \\
\selfautoflow~KITTI (3.2M) &  1.88 & 2.85 & \textbf{4.22} \\

\end{tabularx}
\label{tab:pretrain}
\end{table}

\begin{table}[t]
\centering
\caption{ \textbf{Ablation study on end-to-end training.} The models are trained with the dataset-mixing strategy and longer training steps.}
\vspace{-0.3em}
\label{tab:end2end}
\begin{adjustbox}{width=0.43\textwidth}
\begin{tabular}{lccccc}
& $\mathcal{L}_\text{photo}$ & $\mathcal{L}_{\text{smooth}}$ & $\mathcal{L}_{\text{distill}}$ & $\mathcal{L}_\text{photo}+\mathcal{L}_{\text{smooth}}$ & \bf $\mathcal{L}_\text{total}$\\
\hline
Sintel Clean & 2.26 & 2.18 & 2.07 & 2.30 & \bf 1.83 \\
Sintel Final & 3.24 & 2.84 & 3.04 & 2.98 & \bf 2.59 \\ 
\end{tabular}
\end{adjustbox}
\vspace{-0.3em}
\end{table}

\begin{table*}[t]
    \centering
    \caption{ {\bf Comparison of self-supervised learning approaches.} 
    Our models are pre-trained on self-supervised AutoFlow (\selfautoflow) and the self-supervised objective (SS) using unlabeled data from the target dataset.
    Following SMURF~\cite{stone2021smurf}, we train two models for each dataset on either the training split or the test split and evaluate on the other, denoted as {\bf \selfautoflow+SS train} and {\bf \selfautoflow+SS test}. Our method performs favorably against the state of the art.
    ``\{\}'' trained on/using the unlabeled evaluation set; ``[]'' trained on data closed to evaluation set; ``MF'' using multi-frame estimation at test time~\cite{stone2021smurf}.
    }   
    \resizebox{0.8\textwidth}{!}{
	
    \begin{tabular}{l cc c cc c cccc}
        & \multicolumn{2}{c}{Sintel Clean~\cite{ButlerECCV2012}} & \phantom{aaa} & \multicolumn{2}{c}{Sintel Final~\cite{ButlerECCV2012}} & \phantom{aaa} & \multicolumn{4}{c}{KITTI 2015~\cite{KITTI2015}} \\
         \cmidrule{2-3} \cmidrule{5-6} \cmidrule{8-11}     
        & \multicolumn{2}{c}{AEPE~$\downarrow$} && \multicolumn{2}{c}{AEPE~$\downarrow$} && AEPE~$\downarrow$ & AEPE (noc)~$\downarrow$ & \multicolumn{2}{c}{Fl-all (\%)~$\downarrow$}\\
        Method& \textit{train} & \textit{test} && \textit{train} & \textit{test} && \textit{train} & \textit{train} & \textit{train} & \textit{test}\\
        \midrule
        EPIFlow \cite{Zhong2019UnsupervisedDE} & 3.94 & 7.00 && 5.08 & 8.51 && 5.56 & 2.56 & -- & 16.95 \\ %
        \rowcolor{LightGray}
        UFlow \cite{jonschkowski2020matters} & 3.01 & {5.21} && 4.09 & {6.50} && 2.84 & 1.96 & 9.39 & {11.13} \\
        SemiFlow \cite{im2022semi} & {\bf 1.30} & --& & 2.46 & --& & 3.35  & --& 11.12 & -- \\
        \rowcolor{LightGray}
        SMURF test \cite{stone2021smurf} & 1.99 &  -- &&  2.80 & --  && 2.01 & 1.42 & 6.72 & -- \\ 
        \selfautoflow+SS test & 1.65 & -- && {\bf 2.40} & -- && {\bf 1.94} & {\bf 1.37} & {\bf 6.56} & --\\ \hline
        DDFlow \cite{DDFlow} & \{2.92\} & 6.18 && \{3.98\} & 7.40 && [5.72] & [2.73] & -- & 14.29 \\ %
        \rowcolor{LightGray}
        SelFlow \cite{SelFlow}~$^{\text{(MF)}}$ & [2.88] & [6.56] && \{3.87\} & \{6.57\} && [4.84] & [2.40] & -- & 14.19  \\ %
        UnsupSimFlow \cite{im2020unsupervised} & \{2.86\} & 5.92 && \{3.57\} & 6.92 && [5.19] & -- & -- & [13.38]\\
        \rowcolor{LightGray}
        ARFlow \cite{liu2020learning}~$^{\text{(MF)}}$ & \{2.73\} & \{4.49\} && \{3.69\} & \{5.67\} && [2.85] & -- & -- & [11.79] \\
        RealFlow \cite{han2022realflow}& \{1.34\} & --& & \{2.38\} & --& & \{2.16\} & --& --& -- \\
        \rowcolor{LightGray}
        SMURF train \cite{stone2021smurf} & \{1.71\} & 3.15 && \{2.58\} & 4.18 && \{2.00\} & \{1.41\} & \{6.42\} & 6.83 \\
        \selfautoflow+SS train & \{1.51\} & {\bf 3.03} && \{2.30\} & {\bf 3.98} && \{1.96\} & \{1.38\} & \{6.26\} & {\bf 6.76}\\        
    \end{tabular}
	}
    \label{tab:sota}
\end{table*}

\subsection{Self-supervised AutoFlow}
\label{sec:result1}
\myparagraph{Comparison with the state-of-the-art pre-training approaches.}
\cref{tab:pretrain} compares our method with different pre-training approaches 
and reports the accuracy on Sintel and KITTI.
All methods use RAFT~\cite{RAFT} as the backbone architecture. 
AutoFlow (\autoflow)~\cite{sun2021autoflow} and our Self-AutoFlow (\selfautoflow) are trained on each rendered dataset for Sintel or KITTI, and we report accuracy on both benchmark datasets.
\selfautoflow~mixes rendered datasets from top-3 hyperparameter sets that show low metric score (see \cref{sec:autoflow}); for a fair comparison, we prepare an equivalent model for AutoFlow and denote it as \autoflow-mix. 
``\autoflow~X'', ``\autoflow-mix X'' or ``\selfautoflow~X'' indicates that AutoFlow (\autoflow) or self-supervised AutoFlow (\selfautoflow) is learned for the target domain X.
Our dataset-mixing strategy improves AF-KITTI from their reported number 4.33 to 3.86, demonstrating its effectiveness for both supervised and self-supervised setups.

Our method substantially outperforms (self-)supervised pre-trained models on FlyingChairs and performs competitively to (supervised) AutoFlow and AutoFlow-mix. 
The performance gap between S-AF KITTI and AF KITTI (4.29 \vs 3.86) is much smaller than that between Chairs and AF (7.63 \vs 3.86).
We note that the accuracy in~\cref{tab:pretrain} is reported on the training set, where~\autoflow~uses its ground truth to optimize, and thus is guaranteed to outperform~\selfautoflow. 
It is significant to achieve such a small performance gap, suggesting that our self-supervised approach can successfully extend the applicability of AutoFlow on unlabeled target domains as demonstrated in~\cref{subsec:keypoint_badja}.

\myparagraph{Ablation study of self-supervised search metric.} 
\cref{fig:ablation} provides an ablation study on our search metrics in \cref{eq:search_metric_ours}.
Similar to \cref{fig:correlation}, each data point corresponds to a trained model with its AEPE on Sintel Final ($y$ axis) and a loss value on a metric ($x$ axis) that is used for our \selfautoflow~hyperparameter search to render its training dataset.

Unlike in \cref{fig:correlation} where we observe a strong correlation between the supervised search metric (AEPE) and the measured self-supervised loss, here we observe very different behavior.
Each of the individual self-supervised signals performs  poorly as a search metric, when judged by the AEPE of the models trained on rendered datasets that are searched by the self-supervised signals.
For example, a \selfautoflow~hyperparameter search guided by the distillation loss converges to models with very high AEPE but low distillation loss because distillation alone can lead to trivial solutions, such as a model predicting zero or constant flow for any input.  As a result, only the combination of all three self-supervised signals act as an effective search metric, showing the highest correlation with AEPE and the lowest AEPE ($<3.4$).

\cref{tab:end2end} reports the AEPE of models trained on rendered datasets optimized for different self-supervised metrics. Note, the models in this table use the full training setup, including the dataset-mixing strategy and longer training steps. 
The model with $\mathcal{L}_\text{total}$ shows the lowest AEPE, confirming that the combination of three losses serves as a reliable search metric.

\begin{table}[t]
\centering
\caption{\textbf{Generalization across datasets.} We compare the generalization ability of self-supervised optical flow methods. We train the models on one dataset and evaluate on others. Our method (\selfautoflow) outperforms SMURF on cross-dataset evaluations.
SS Sintel/KITTI means further self-supervised training on Sintel/KITTI.
}
\resizebox{0.95\linewidth}{!}{
\centering
\begin{tabular}{lccccccccc}
&\phantom{a}& Chairs & \phantom{a} & \multicolumn{2}{c}{Sintel \textit{train}} & \phantom{a} & \multicolumn{2}{c}{KITTI-15 \textit{train}} \\
\cmidrule{3-3} \cmidrule{5-6} \cmidrule{8-9} 
Method && \textit{test} && Clean & Final && AEPE & Fl-all (\%)\\
\midrule
SMURF Chairs &&  1.72 &&  2.19 & 3.35 && 7.94 &  26.51 \\
\selfautoflow~Sintel && \bf 1.61 && \bf 1.83 & \bf 2.57 && 4.79 & 15.47 \\
\selfautoflow~KITTI && 2.09 && 2.16 & 2.96 && \bf 4.28 & \bf 13.60 \\
\midrule
{+ SS Sintel}\\
SMURF && 1.99 &&  1.99 & 2.80 && 4.47 & 12.55 \\
\selfautoflow  && \bf 1.81 &&  \bf 1.65 &  \bf 2.40 && \bf 4.28 & \bf 12.45 \\
\midrule
{+ SS KITTI}\\
SMURF  && 3.26 &&  3.38 &  4.47 &&  2.01 &  6.72 \\
\selfautoflow  && \bf 3.19 && \bf 3.32 & \bf 4.44 &&  \bf 1.94 &  \bf 6.56 \\
\end{tabular}
}
\label{tab:general}
\end{table}

\subsection{Self-supervised Learning of Optical Flow}
\myparagraph{Comparison to the state of the art.}
In \cref{sec:combine}, we combine our \selfautoflow~(\cref{sec:autoflow}) with the self-supervised learning approach of optical flow to further adapt the model to the target domains, denoted by {\bf S-AF+SS}.
We compare against state-of-the-art approaches that do not use ground truth in the target domain in~\cref{tab:sota}.
We train our model on the standard train/test splits for Sintel and further train on the multi-view extension data following~\cite{stone2021smurf} for the KITTI dataset.
We train two models for each dataset, one trained on the test split (* test) in a self-supervised manner and evaluated on the training split with ground truth, and the other trained on the training split (* train) and evaluated on the test split (\ie, benchmark websites).

Compared to SMURF, our method reduces the AEPE by 0.12 on Sintel Clean test, 0.20 on Sintel Final test, and F1-all by 0.07 on KITTI test. Our method is comparable to SemiFlow \cite{im2022semi} and RealFlow \cite{han2022realflow} on Sintel Clean train, although both SemiFlow and RealFlow are pre-trained on FlyingChairs and FlyingThings3D and thus have strong performance on Sintel Clean, due to the proximity of their domains. 
Our method outperforms SemiFlow and RealFlow on the more challenging Sintel Final train and KITTI.

\begin{figure*}[t]
    \centering
    \includegraphics[width=1.0\textwidth]{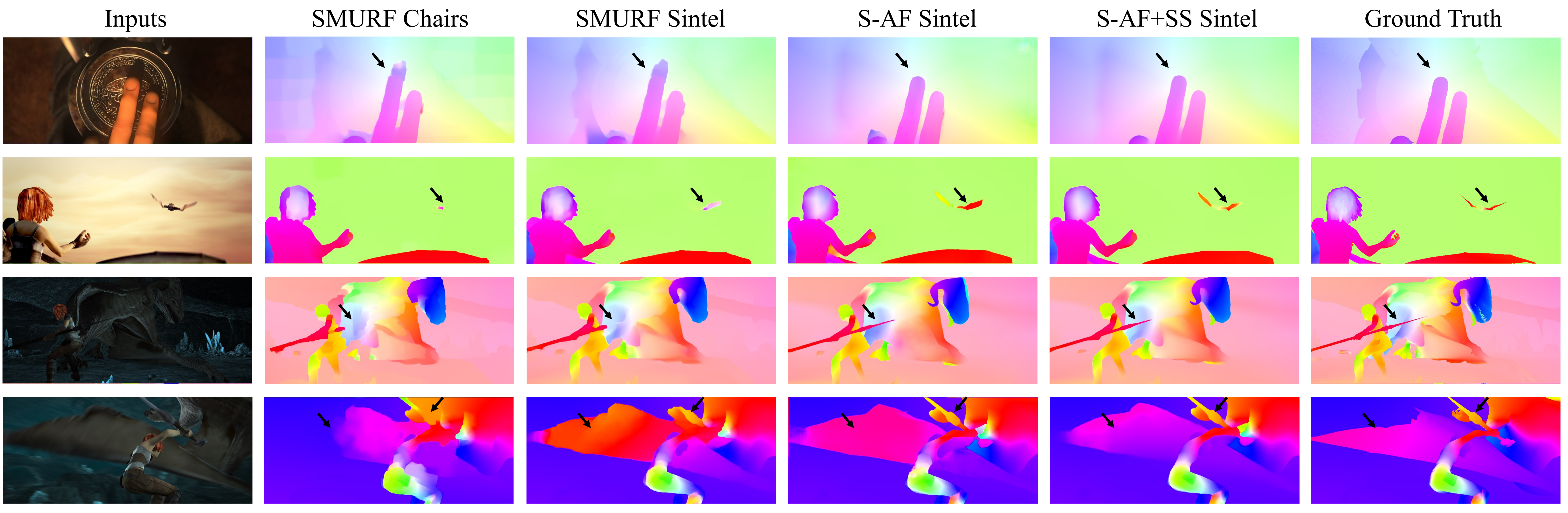}
    \caption{
    \textbf{Comparison of self-supervised methods on Sintel.} SMURF, both pre-trained (SMURF Chairs) and self-supervised fine-tuned (SMURF Sintel), tends to fail on shadows, strong motion blur, or small/thin objects.
    On the other hand, self-supervised AutoFlow (\selfautoflow) on Sintel provides more reliable predictions, and self-supervised (SS) fine-tuning (\selfautoflow+SS Sintel) further improves the results.
    }
    \label{fig:sintel}
\end{figure*}

\myparagraph{Generalization across datasets.}
In \cref{tab:general}, we evaluate the generalization of our approach by training the model on one dataset and evaluating it on other datasets.
We denote the models with self-supervised fine-tuning on target datasets as +SS Sintel/KITTI.
When only training on \selfautoflow~datasets, the model achieves an AEPE of 1.83 on Sintel Clean and 2.57 on Sintel Final, which outperforms SMURF with self-supervised fine-tuning on the target Sintel dataset by 0.16 and 0.23.
Both models trained on \selfautoflow~and self-supervised fine-tuned on Sintel/KITTI achieve the best cross-domain performance on all the target datasets.

\subsection{Supervised Fine-tuning on Public Benchmarks}
To examine how well our method can serve as a good initialization, we fine-tune our \selfautoflow+SS train model in~\cref{tab:sota} using the same fine-tuning protocol from~\cite{sun2022disentangling}.
As shown in \cref{tab:benchmark}, our method consistently outperforms RAFT-it~\cite{sun2022disentangling}, SemiFlow, and RealFlow, indicating that \selfautoflow+SS models can serve as a good initialization for supervised fine-tuning. 

\begin{table}[t]
\centering
\footnotesize
\caption{\textbf{Supervised fine-tuning on public benchmarks.}
We fine-tune our model using ground truth in a supervised manner. (AEPE~$\downarrow$ for Sintel and Fl-all~$\downarrow$ for KITTI. Methods using warm start on Sintel are marked by *). Models pre-trained on self-supervised AutoFlow (\selfautoflow) can serve as a good initialization for supervised fine-tuning.}
\label{tab:benchmark}
\begin{tabular}{lccc} 
Method & Sintel Clean & Sintel Final & KITTI \\ \hline
RealFlow \cite{han2022realflow} & - & - & 4.63 \% \\
SemiFlow (RAFT)* \cite{im2022semi} & 1.65 & 2.79 & 4.85 \% \\
RAFT-it~\cite{sun2022disentangling}   & {1.55} & {2.90}& {4.31 \%} \\ 
RAFT-\selfautoflow & \bf 1.42 & \bf 2.75 & \bf 4.12 \% \\
\end{tabular}
\end{table}

\subsection{Evaluation on Downstream Tasks}
\label{subsec:keypoint_badja}
To further demonstrate the generalization of our method to a real-world domain without ground truth, we compare our method with various supervised, semi-supervised, and self-supervised methods on two downstream tasks: keypoint propagation and segmentation tracking on the DAVIS dataset \cite{DAVIS}. 
For self-supervised fine-tuning on DAVIS, we use the seven BADJA sequences and three challenging sequences (drift-turn, drift-chicane and color-run) as the test set, and the remaining 80 sequences for training.

\myparagraph{Keypoint propagation.}
For evaluation, we use the Percentage of Correct Keypoint-Transfer (PCK-T) metric~\cite{yang2012articulated} with keypoint annotations from the BADJA dataset~\cite{biggs2018creatures}.
Given annotated keypoints on a reference image, the metric calculates the percentage of correctly propagated keypoints along a video sequence. 
As shown in \cref{tab:badja_test}, our \selfautoflow~DAVIS model achieves better accuracy than other (semi-)supervised approaches (SemiFlow, RealFlow, \autoflow~Sintel, and RAFT-it) and a self-supervised pre-training approach (SMURF Chairs). 
Our self-supervised fine-tuned model on DAVIS (\selfautoflow+SS DAVIS) outperforms SMURF DAVIS.

Compared to the \selfautoflow~results that use different unlabeled data as target, \selfautoflow~DAVIS outperforms \selfautoflow~Sintel and KITTI, showing that our method successfully learns a better dataset for the target domains without using the ground truth labels.

\myparagraph{Segmentation tracking.}
We propagate initial segmentation masks using optical flow and evaluate IoU between the propagated and ground truth masks. 
As shown in \cref{tab:segmentation_tracking}, Our method (S-AF) consistently outperforms supervised AutoFlow, RAFT, and SMURF.
Since the performance difference is mainly on tiny objects or around object boundaries, the $>1\%$ difference between S-AF+SS Davis and SMURF Davis is a moderate improvement.

\begin{table}[t]
\centering
\caption{\textbf{Keypoint propagation on the BADJA dataset \cite{biggs2018creatures}.} We use different optical flow methods to propagate the keypoints along the sequences and report the PCK-T metric. 
(S-)AF: (self-supervised) AutoFlow; SS: self-supervised fine-tuning. SMURF DAVIS is first trained on Chairs and then fine-tuned on DAVIS.
}
\label{tab:badja_test}
\begin{adjustbox}{width=0.47\textwidth}
\begin{tabular}{lccccccc|c}
{Method} & bear & camel & cows & dog-a & dog & horse-h & horse-l & \bf Avg. \\ \hline
DINO \cite{caron2021emerging} & 75.7 & 58.2 & 71.4 & 10.3 & 46.0 & 35.8 & 56.5 & 50.6 \\
PIPs \cite{harley2022particle} & 76.3 & 84.0 & 79.1 & 31.6 & 42.9 & 60.4 & 58.6 & \bf 61.8 \\ \hline
\bf (Semi-)supervised \\
SemiFlow-Davis \cite{im2022semi} & 66.4 & 72.0 & 71.4 & 13.8 & 40.8 & 36.4 & 31.4 & 47.5 \\
RealFlow-Davis \cite{han2022realflow} & 64.3 & 80.1 & 63.4 & 10.3 & 45.4 & 32.5 & 38.7 & 47.8 \\
\autoflow~Sintel \cite{sun2021autoflow} & 71.4 & 80.1 & 75.1 & 17.2 & 47.1 & 34.4 & 27.2 & 50.4 \\
RAFT-it \cite{sun2022disentangling} & 73.2 & 83.0 & 78.1 & 17.2 & 46.0 & 39.1 & 30.4 & \bf 52.4 \\ \hline

\bf Pre-training \\ 
SMURF Chairs~\cite{stone2021smurf} & 79.3 & 74.0 & 73.8 & 3.4 & 42.5 & 34.4 & 29.3 & 48.1 \\
\selfautoflow~Sintel & 73.2 & 83.9 & 62.0 & 3.4 & 42.0 & 40.4 & 26.7 & 47.4 \\
\selfautoflow~KITTI & 72.5 & 76.8 & 73.8 & 0.0 & 46.6 & 34.4 & 31.9 & 48.0 \\
\selfautoflow~DAVIS & 72.9 & 76.5 & 75.7 & 20.7 & 47.7 & 38.4 & 31.4 & \bf 51.9 \\ \hline
\bf Self-supervised fine-tuning \\ 
SMURF DAVIS \cite{stone2021smurf} & 80.0 & 83.0 & 77.8 & 3.4 & 47.1 & 40.4 & 44.0 & 53.7 \\
\selfautoflow+SS DAVIS & 80.0 & 82.3 & 74.9 & 10.3 & 50.6 & 43.0 & 42.4 & \bf 54.8 \\
\end{tabular}
\end{adjustbox}
\end{table}

\begin{table}[t]
\centering
\caption{\textbf{Segmentation tracking on DAVIS.} We propagate the initial segmentation masks using optical flow and evaluate IoU compared to ground truth masks. }
\begin{adjustbox}{width=0.48\textwidth}
\begin{tabular}{cccc|cc} 
AF Sintel & RAFT-it & SMURF Chairs & \bf S-AF Davis & SMURF Davis & \bf S-AF+SS Davis
\\ \hline
0.830 & 0.801 & 0.807 & \bf 0.837 & 0.876 & \bf 0.888\\
\end{tabular}
\label{tab:segmentation_tracking}
\end{adjustbox}
\end{table}

\begin{figure*}[t]
    \centering
    \includegraphics[width=1.0\textwidth]{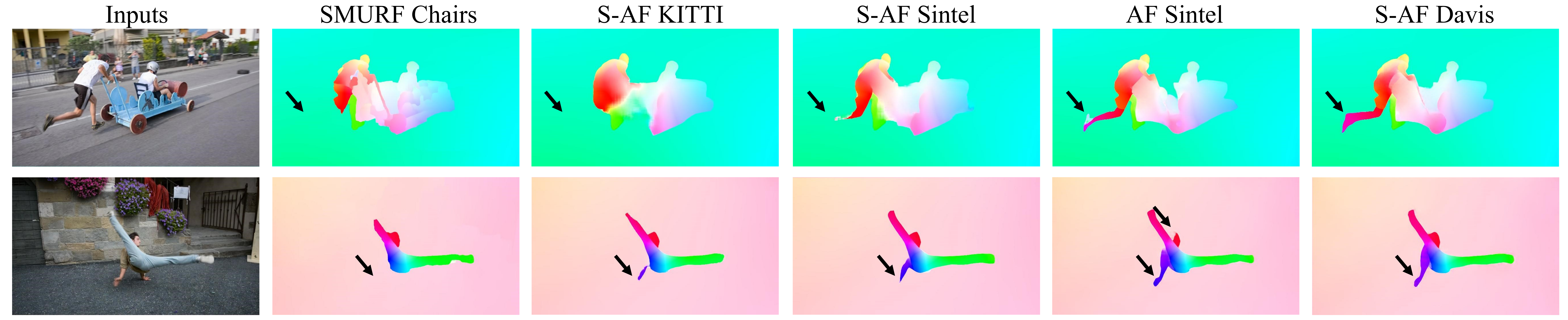}
    \caption{\textbf{Visual comparison of pre-training on DAVIS.} 
    Compared to SMURF Chairs and AutoFlow (AF) Sintel, self-supervised AutoFlow (\selfautoflow) DAVIS learned from DAVIS data yields better flow results. 
    In addition, \selfautoflow~DAVIS outperforms \selfautoflow~Sintel and \selfautoflow~KITTI, indicating \selfautoflow~successfully learns a better training set to adapt the model to a target domain. 
    }
    \label{fig:davis_pretraining}
\end{figure*}

\begin{figure}[t]
    \centering
    \includegraphics[width=\columnwidth]{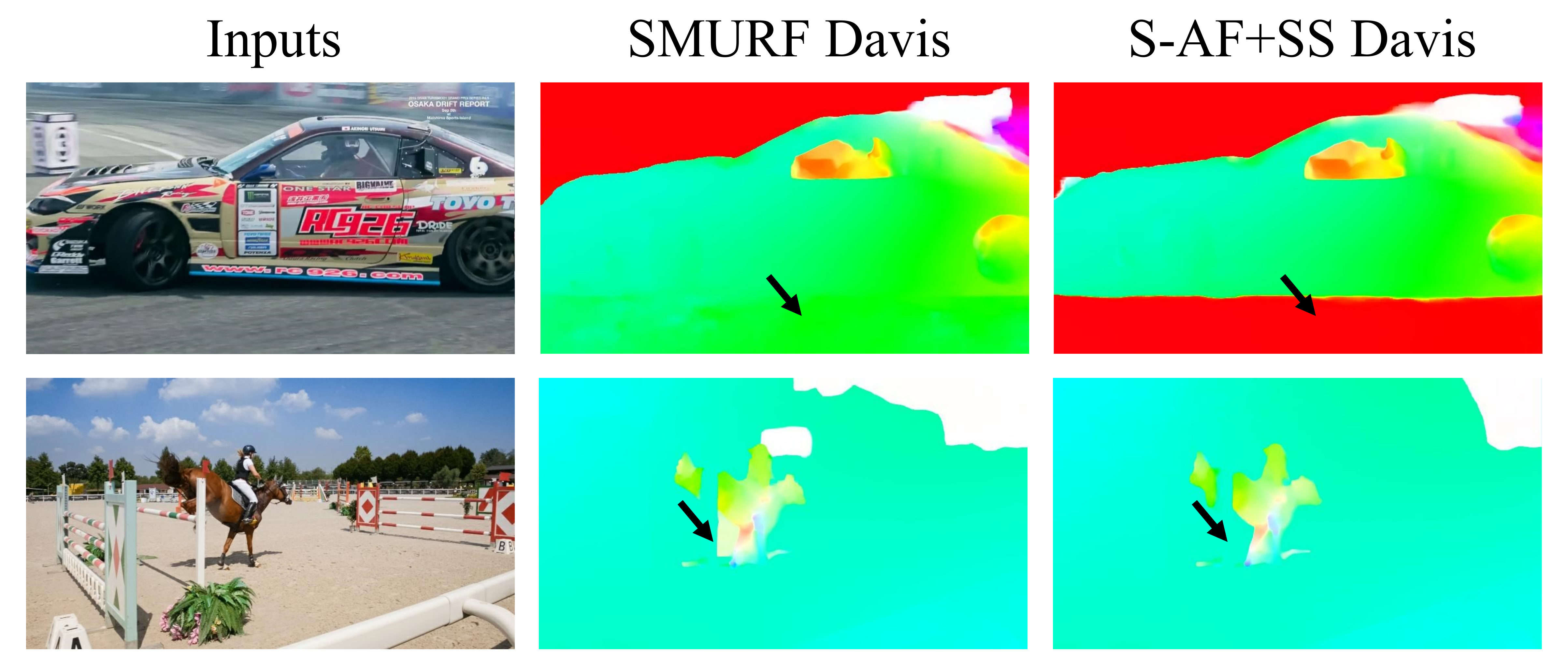}
    \caption{\textbf{Visual comparison of self-supervised fine-tuning on DAVIS.} With self-supervised fine-tuning on the target DAVIS dataset, our \selfautoflow+SS DAVIS predicts more accurate flow for textureless areas or thin objects, showing that the better initialization of \selfautoflow~leads to better self-supervised fine-tuning results compared to the initialization from pre-training on FlyingChairs.
    } 
    \label{fig:davis_ours_vs_smurf}
    \vspace{-0.5em}
\end{figure}

\begin{figure}[t]
    \centering
    \includegraphics[width=\columnwidth]{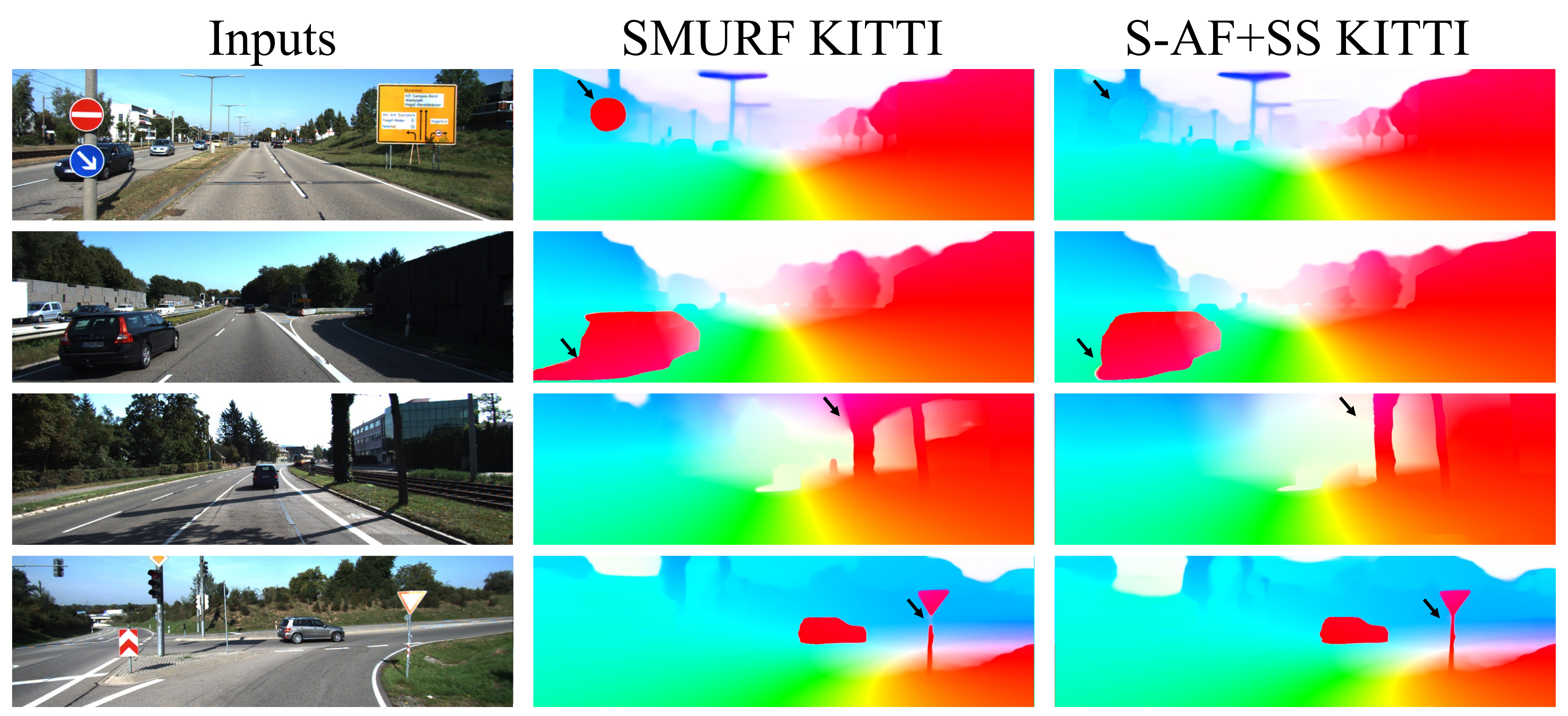}
    \caption{\textbf{Visual comparison of self-supervised fine-tuning on KITTI.} Our \selfautoflow+SS KITTI model predicts more accurate flow on objects with large motion, on shadows, and on thin structures of the scene compared to SMURF KITTI. Purely self-supervised methods may predict incorrect flow fields, while our \selfautoflow+SS approach resolves this issue by a better initialization.
    }
    \label{fig:kitti}
\end{figure}

\subsection{Visual Comparison}
\myparagraph{Sintel.}
As in~\cref{fig:sintel}, compared to out-of-domain pre-training approaches, \selfautoflow~Sintel performs better than SMURF Chairs on shadows, small/thin objects and scenes with strong motion blur.
Self-supervised fine-tuning on Sintel (\selfautoflow+SS Sintel model) further improves the results upon the pre-training \selfautoflow~Sintel model, whereas SMURF still tends to fail on those cases.
The results show that our self-supervised learning-to-render approach not only provides a strong pre-trained model on the target domain, but also serves as a good initialization for the self-supervised fine-tuning; suggesting that our \selfautoflow~is complementary to the self-supervised learning approach.

\myparagraph{DAVIS.}
\cref{fig:davis_pretraining} shows the comparison of different pre-training methods on the DAVIS dataset. 
The SMURF Chairs model does not clearly capture the motion of the foot and hand of the person due to the domain gap between FlyingChairs and DAVIS. 
\autoflow~Sintel does not generalize well to the real-world DAVIS data; our \selfautoflow~learned from the unlabeled DAVIS data successfully captures the detailed structure.
We further compare our models that use different target domains for dataset generation (\selfautoflow~DAVIS, \selfautoflow~Sintel, \selfautoflow~KITTI). 
\selfautoflow~DAVIS shows the best results by successfully optimizing the rendering parameters for the real-world target domain, \ie DAVIS.

\cref{fig:davis_ours_vs_smurf} shows that self-supervised fine-tuning on DAVIS (\selfautoflow+SS DAVIS) further improves the result over SMURF DAVIS, showing that the better initialization of \selfautoflow~leads to better self-supervised fine-tuning results compared to the initialization from pre-training on FlyingChairs.

\myparagraph{KITTI.} 
As shown in~\cref{fig:kitti}, \selfautoflow+SS KITTI predicts more accurate flow on close objects with large motion,  shadows, and thin structures of the scene than SMURF KITTI.
The results suggest that the purely self-supervised method may predict incorrect flow due to optimizing the photometric constancy loss. 
In contrast, our \selfautoflow~pre-training approach provides a better model initialization and resolves this issue by pre-training on rendered \selfautoflow~data ground truth.

\myparagraph{Discussions.}
Despite the promising results, the visual comparison suggests that there is room for improvement, such as the thin structures in~\cref{fig:davis_pretraining} and the sky regions in~\cref{fig:davis_ours_vs_smurf}. Future work may further explore using a more realistic rendering engine \eg, with a sky model, and developing better self-supervised losses to address these issues.

\section{Conclusions}
We have introduced self-supervised AutoFlow to learn a training set for optical flow for unlabeled data using self-supervised metrics. 
Self-supervised AutoFlow performs on par with AutoFlow that uses ground truth on Sintel and KITTI, and better on the real-world DAVIS dataset where ground truth is not available. 
Our work suggests the benefits of connecting learning to render with self-supervision and we hope to see more work in this direction to solve optical flow in the real world. 
%

{
\small
\bibliographystyle{ieee_fullname}
\bibliography{flow}

\begin{thebibliography}{10}\itemsep=-1pt

\bibitem{ahmadi2016unsupervised}
Aria Ahmadi and Ioannis Patras.
\newblock Unsupervised convolutional neural networks for motion estimation.
\newblock In {\em {ICIP}}, 2016.

\bibitem{aleotti2021learning}
Filippo Aleotti, Matteo Poggi, and Stefano Mattoccia.
\newblock Learning optical flow from still images.
\newblock In {\em {CVPR}}, 2021.

\bibitem{biggs2018creatures}
Benjamin Biggs, Thomas Roddick, Andrew Fitzgibbon, and Roberto Cipolla.
\newblock Creatures great and {SMAL}: {R}ecovering the shape and motion of
  animals from video.
\newblock In {\em {ACCV}}, 2018.

\bibitem{ButlerECCV2012}
Daniel~J. Butler, Jonas Wulff, Garrett~B. Stanley, and Michael~J. Black.
\newblock A naturalistic open source movie for optical flow evaluation.
\newblock In {\em {ECCV}}, 2012.

\bibitem{caron2021emerging}
Mathilde Caron, Hugo Touvron, Ishan Misra, Herv{\'e} J{\'e}gou, Julien Mairal,
  Piotr Bojanowski, and Armand Joulin.
\newblock Emerging properties in self-supervised vision transformers.
\newblock In {\em {ICCV}}, 2021.

\bibitem{FlowNet}
Alexey Dosovitskiy, Philipp Fischer, Eddy Ilg, Philip H{\"a}usser, Caner
  Haz{\i}rba{\c{s}}, Vladimir Golkov, Patrick {van der Smagt}, Daniel Cremers,
  and Thomas Brox.
\newblock Flow{N}et: {L}earning optical flow with convolutional networks.
\newblock In {\em {ICCV}}, 2015.

\bibitem{goodfellow2014generative}
Ian~J. Goodfellow, Jean Pouget-Abadie, Mehdi Mirza, Bing Xu, David
  Warde-Farley, Sherjil Ozair, Aaron Courville, and Yoshua Bengio.
\newblock Generative adversarial nets.
\newblock In {\em {NIPS}}, 2014.

\bibitem{greff2022kubric}
Klaus Greff, Francois Belletti, Lucas Beyer, Carl Doersch, Yilun Du, Daniel
  Duckworth, David~J. Fleet, Dan Gnanapragasam, Florian Golemo, Charles
  Herrmann, et~al.
\newblock Kubric: A scalable dataset generator.
\newblock In {\em {CVPR}}, 2022.

\bibitem{han2022realflow}
Yunhui Han, Kunming Luo, Ao Luo, Jiangyu Liu, Haoqiang Fan, Guiming Luo, and
  Shuaicheng Liu.
\newblock Real{F}low: {EM}-based realistic optical flow dataset generation from
  videos.
\newblock In {\em {ECCV}}, 2022.

\bibitem{harley2022particle}
Adam~W. Harley, Zhaoyuan Fang, and Katerina Fragkiadaki.
\newblock Particle video revisited: {T}racking through occlusions using point
  trajectories.
\newblock In {\em {ECCV}}, 2022.

\bibitem{huang2022flowformer}
Zhaoyang Huang, Xiaoyu Shi, Chao Zhang, Qiang Wang, Ka~Chun Cheung, Hongwei
  Qin, Jifeng Dai, and Hongsheng Li.
\newblock Flow{F}ormer: {A} transformer architecture for optical flow.
\newblock In {\em {ECCV}}, 2022.

\bibitem{Hui_2018_CVPR}
Tak-Wai Hui, Xiaoou Tang, and Chen Change~Loy.
\newblock Lite{F}low{N}et: {A} lightweight convolutional neural network for
  optical flow estimation.
\newblock In {\em {CVPR}}, 2018.

\bibitem{humby2006data}
Clive Humby.
\newblock Data is the new oil.
\newblock {\em Proc. ANA Sr. Marketer’s Summit. Evanston, IL, USA}, 2006.

\bibitem{Flownet2}
Eddy Ilg, Nikolaus Mayer, Tonmoy Saikia, Margret Keuper, Alexey Dosovitskiy,
  and Thomas Brox.
\newblock {FlowNet} 2.0: {E}volution of optical flow estimation with deep
  networks.
\newblock In {\em {CVPR}}, 2017.

\bibitem{im2020unsupervised}
Woobin Im, Tae-Kyun Kim, and Sung-Eui Yoon.
\newblock Unsupervised learning of optical flow with deep feature similarity.
\newblock In {\em {ECCV}}, 2020.

\bibitem{im2022semi}
Woobin Im, Sebin Lee, and Sung-Eui Yoon.
\newblock Semi-supervised learning of optical flow by flow supervisor.
\newblock In {\em {ECCV}}, 2022.

\bibitem{jaegle2021perceiver}
Andrew Jaegle, Sebastian Borgeaud, Jean-Baptiste Alayrac, Carl Doersch, Catalin
  Ionescu, David Ding, Skanda Koppula, Daniel Zoran, Andrew Brock, Evan
  Shelhamer, et~al.
\newblock Perceiver {IO}: A general architecture for structured inputs \&
  outputs.
\newblock In {\em {ICLR}}, 2022.

\bibitem{jahedi2022multi}
Azin Jahedi, Lukas Mehl, Marc Rivinius, and Andr{\'e}s Bruhn.
\newblock Multi-{S}cale {RAFT}: {C}ombining hierarchical concepts for
  learning-based optical flow estimation.
\newblock In {\em {ICIP}}, 2022.

\bibitem{jiang2021learning}
Shihao Jiang, Dylan Campbell, Yao Lu, Hongdong Li, and Richard Hartley.
\newblock Learning to estimate hidden motions with global motion aggregation.
\newblock In {\em {ICCV}}, 2021.

\bibitem{jonschkowski2020matters}
Rico Jonschkowski, Austin Stone, Jonathan~T. Barron, Ariel Gordon, Kurt
  Konolige, and Anelia Angelova.
\newblock What matters in unsupervised optical flow.
\newblock In {\em {ECCV}}, 2020.

\bibitem{Alexnet}
Alex Krizhevsky, Ilya Sutskever, and Geoffrey~E. Hinton.
\newblock Image{N}et classification with deep convolutional neural networks.
\newblock In {\em {NIPS}}, 2012.

\bibitem{lai2017semi}
Wei-Sheng Lai, Jia-Bin Huang, and Ming-Hsuan Yang.
\newblock Semi-supervised learning for optical flow with generative adversarial
  networks.
\newblock In {\em {NIPS}}, 2017.

\bibitem{liu2020learning}
Liang Liu, Jiangning Zhang, Ruifei He, Yong Liu, Yabiao Wang, Ying Tai, Donghao
  Luo, Chengjie Wang, Jilin Li, and Feiyue Huang.
\newblock Learning by {A}nalogy: {R}eliable supervision from transformations
  for unsupervised optical flow estimation.
\newblock In {\em {CVPR}}, 2020.

\bibitem{DDFlow}
Pengpeng Liu, Irwin King, Michael~R. Lyu, and Jia Xu.
\newblock {DDF}low: Learning optical flow with unlabeled data distillation.
\newblock In {\em AAAI}, 2019.

\bibitem{SelFlow}
Pengpeng Liu, Michael~R. Lyu, Irwin King, and Jia Xu.
\newblock Sel{F}low: {S}elf-supervised learning of optical flow.
\newblock In {\em {CVPR}}, 2019.

\bibitem{Mayer2018}
Nikolaus Mayer, Eddy Ilg, Philipp Fischer, Caner Hazirbas, Daniel Cremers,
  Alexey Dosovitskiy, and Thomas Brox.
\newblock What makes good synthetic training data for learning disparity and
  optical flow estimation?
\newblock {\em IJCV}, 2018.

\bibitem{meister2018unflow}
Simon Meister, Junhwa Hur, and Stefan Roth.
\newblock Un{F}low: {U}nsupervised learning of optical flow with a
  bidirectional census loss.
\newblock In {\em AAAI}, 2018.

\bibitem{KITTI2015}
Moritz Menze, Christian Heipke, and Andreas Geiger.
\newblock Joint 3{D} estimation of vehicles and scene flow.
\newblock In {\em ISPRS Workshop on Image Sequence Analysis (ISA)}, 2015.

\bibitem{DAVIS}
Jordi Pont-Tuset, Federico Perazzi, Sergi Caelles, Pablo Arbel\'aez, Alexander
  Sorkine-Hornung, and Luc {Van Gool}.
\newblock The 2017 {DAVIS} challenge on video object segmentation.
\newblock {\em arXiv preprint arXiv:1704.00675}, 2017.

\bibitem{ranftl2021vision}
Ren{\'e} Ranftl, Alexey Bochkovskiy, and Vladlen Koltun.
\newblock Vision transformers for dense prediction.
\newblock In {\em {ICCV}}, 2021.

\bibitem{ranftl2020towards}
Ren{\'e} Ranftl, Katrin Lasinger, David Hafner, Konrad Schindler, and Vladlen
  Koltun.
\newblock Towards robust monocular depth estimation: Mixing datasets for
  zero-shot cross-dataset transfer.
\newblock {\em {PAMI}}, 2020.

\bibitem{spynet2017}
Anurag Ranjan and Michael~J. Black.
\newblock Optical flow estimation using a spatial pyramid network.
\newblock In {\em CVPR}, 2017.

\bibitem{ren2017unsupervised}
Zhe Ren, Junchi Yan, Bingbing Ni, Bin Liu, Xiaokang Yang, and Hongyuan Zha.
\newblock Unsupervised deep learning for optical flow estimation.
\newblock In {\em AAAI}, 2017.

\bibitem{richter2017playing}
Stephan~R. Richter, Zeeshan Hayder, and Vladlen Koltun.
\newblock Playing for benchmarks.
\newblock In {\em {ICCV}}, 2017.

\bibitem{ronneberger2015u}
Olaf Ronneberger, Philipp Fischer, and Thomas Brox.
\newblock U-{N}et: {C}onvolutional networks for biomedical image segmentation.
\newblock In {\em MICCAI}, 2015.

\bibitem{russakovsky2015imagenet}
Olga Russakovsky, Jia Deng, Hao Su, Jonathan Krause, Sanjeev Satheesh, Sean Ma,
  Zhiheng Huang, Andrej Karpathy, Aditya Khosla, Michael Bernstein, et~al.
\newblock Image{N}et large scale visual recognition challenge.
\newblock {\em {IJCV}}, 115(3):211--252, 2015.

\bibitem{smith2019super}
Leslie~N. Smith and Nicholay Topin.
\newblock Super-convergence: Very fast training of neural networks using large
  learning rates.
\newblock In {\em Artificial intelligence and machine learning for multi-domain
  operations applications}, volume 11006, pages 369--386. SPIE, 2019.

\bibitem{stone2021smurf}
Austin Stone, Daniel Maurer, Alper Ayvaci, Anelia Angelova, and Rico
  Jonschkowski.
\newblock {SMURF}: Self-teaching multi-frame unsupervised raft with full-image
  warping.
\newblock In {\em {CVPR}}, 2021.

\bibitem{sui2022craft}
Xiuchao Sui, Shaohua Li, Xue Geng, Yan Wu, Xinxing Xu, Yong Liu, Rick Goh, and
  Hongyuan Zhu.
\newblock {CRAFT}: {C}ross-attentional flow transformer for robust optical
  flow.
\newblock In {\em {CVPR}}, 2022.

\bibitem{sun2022disentangling}
Deqing Sun, Charles Herrmann, Fitsum Reda, Michael Rubinstein, David~J. Fleet,
  and William~T. Freeman.
\newblock Disentangling architecture and training for optical flow.
\newblock In {\em {ECCV}}, 2022.

\bibitem{sun2021autoflow}
Deqing Sun, Daniel Vlasic, Charles Herrmann, Varun Jampani, Michael Krainin,
  Huiwen Chang, Ramin Zabih, William~T. Freeman, and Ce Liu.
\newblock Auto{F}low: {L}earning a better training set for optical flow.
\newblock In {\em {CVPR}}, 2021.

\bibitem{Sun2018PWCNet}
Deqing Sun, Xiaodong Yang, Ming-Yu Liu, and Jan Kautz.
\newblock {PWC-Net}: {CNN}s for optical flow using pyramid, warping, and cost
  volume.
\newblock In {\em {CVPR}}, 2018.

\bibitem{RAFT}
Zachary Teed and Jia Deng.
\newblock {RAFT}: Recurrent all-pairs field transforms for optical flow.
\newblock In {\em {ECCV}}, 2020.

\bibitem{wang2018occlusion}
Yang Wang, Yi Yang, Zhenheng Yang, Liang Zhao, Peng Wang, and Wei Xu.
\newblock Occlusion aware unsupervised learning of optical flow.
\newblock In {\em {CVPR}}, 2018.

\bibitem{xiao2020learnable}
Taihong Xiao, Jinwei Yuan, Deqing Sun, Qifei Wang, Xin-Yu Zhang, Kehan Xu, and
  Ming-Hsuan Yang.
\newblock Learnable cost volume using the cayley representation.
\newblock In {\em {ECCV}}, 2020.

\bibitem{xu2021flow1d}
Haofei Xu, Jiaolong Yang, Jianfei Cai, Juyong Zhang, and Xin Tong.
\newblock High-resolution optical flow from 1{D} attention and correlation.
\newblock In {\em {ICCV}}, 2021.

\bibitem{yang2012articulated}
Yi Yang and Deva Ramanan.
\newblock Articulated human detection with flexible mixtures of parts.
\newblock {\em {PAMI}}, 35(12):2878--2890, 2012.

\bibitem{yang2018conditional}
Yanchao Yang and Stefano Soatto.
\newblock Conditional prior networks for optical flow.
\newblock In {\em {ECCV}}, 2018.

\bibitem{jason2016back2basics}
Jason~J. Yu, Adam~W. Harley, and Konstantinos~G. Derpanis.
\newblock Back to {B}asics: {U}nsupervised learning of optical flow via
  brightness constancy and motion smoothness.
\newblock In {\em ECCVW}, 2016.

\bibitem{yuan2022optical}
Shuai Yuan, Xian Sun, Hannah Kim, Shuzhi Yu, and Carlo Tomasi.
\newblock Optical flow training under limited label budget via active learning.
\newblock In {\em {ECCV}}, 2022.

\bibitem{zabih1994non}
Ramin Zabih and John Woodfill.
\newblock Non-parametric local transforms for computing visual correspondence.
\newblock In {\em {ECCV}}, 1994.

\bibitem{zhang2021separable}
Feihu Zhang, Oliver~J. Woodford, Victor~Adrian Prisacariu, and Philip~H.S.
  Torr.
\newblock Separable flow: Learning motion cost volumes for optical flow
  estimation.
\newblock In {\em {ICCV}}, 2021.

\bibitem{Zhong2019UnsupervisedDE}
Yiran Zhong, Pan Ji, Jianyuan Wang, Yuchao Dai, and Hongdong Li.
\newblock Unsupervised deep epipolar flow for stationary or dynamic scenes.
\newblock In {\em {CVPR}}, 2019.

\end{thebibliography}
}

\onecolumn
\maketitle

\section*{Appendix}
\appendix

We first discuss implementation and experiment details.
Next, we present the ablation studies of our approach.
Third, we provide additional analysis of the proposed design.
Finally, we include more visual results.

\section{Implementation Details}
\label{supp:more_details_implementation}

\myparagraph{Training details.}
We use \SI{80}{\kilo{}} training steps for the rendering hyperparameter search.
We include the following rendering hyperparameters for generating the S-AF data:
\begin{itemize}
    \item Number of foreground objects 
    \item Scale, rotation, translation, grid strength, grid size of the motion for foreground
    \item Scale, rotation, translation, grid strength, grid size of the motion for background
    \item Probability and strength of the mask blur
    \item Probability and strength of the motion blur
    \item Probability, density and brightness of the fog
    \item Minimum and maximum of the object’s diagonal
    \item Minimum and maximum of the object’s center location
    \item Irregularity and spikiness of the polygon
\end{itemize}

As for the hyperparameters in the search metric \cref{eq:search_metric_ours} in the main paper, we use $(w_\text{smooth}, w_\text{distill}){=}(0.6, 4)$ for the Sintel \cite{ButlerECCV2012} and the DAVIS dataset \cite{DAVIS}, and $(w_\text{smooth}, w_\text{distill}){=}(1.2, 8)$ for the KITTI dataset \cite{KITTI2015}. 
We pretrain the model on a generated \selfautoflow~dataset $\mathbf{D}_\text{auto}$ for \SI[round-precision=1]{3.2}{\mega{}} iterations for Sintel and \SI{200}{\kilo{}} iterations for KITTI and DAVIS. 
We randomly crop input images to size 368$\times$496 at training time and use a batch size of 36.

We further fine-tune the model with the self-supervised loss (\cref{eq:ft_self_sup_loss}) for \SI{12}{\kilo{}} iterations on the Sintel dataset, \SI{75}{\kilo{}} iterations on the KITTI dataset, and \SI{100}{\kilo{}} iterations on the Davis dataset. 
We further apply the multi-frame fine-tuning on Sintel and KITTI datasets for \SI{30}{\kilo{}} iterations~(\cref{eq:ft_multiframe}) with the same parameter setting from SMURF \cite{stone2021smurf}.
We randomly crop input images to size 368$\times$496 at training time and use a batch size of 8.
We use the data augmentations from RAFT~\cite{RAFT} including random cropping, stretching, scaling, flipping, and erasing. 
As for the photometric augmentations, we randomly adjust the contrast, saturation, brightness and hue.

\myparagraph{Evaluation metrics.}
We use the average end-point error (AEPE) evaluation metric.
For KITTI, we additionally report the outlier rate (Fl-all), \ie~the ratio (in \%) of outlier pixels among all ground truth pixels. 
If an error of a pixel exceeds the 3-pixel threshold and 5\% \wrt the ground truth, the pixel is considered as an outlier.

\section{Ablation Studies}

\subsection{Training by individual S-AF dataset and mixed S-AF dataset}

As described in~\cref{sec:autoflow} and~\cref{sec:result1}, to improve the robustness of the algorithm, we sort the sets of hyperparameters returned by Self-AutoFlow according to the self-supervised search metric and choose the top-3 hyperparameter sets. 
We form our final Self-AutoFlow dataset by equally mixing a set of images generated from each hyperparameter set. For a fair comparison, we also prepare an equivalent model for AutoFlow, denoted as AF-mix.
In addition to the results of training on the dataset generated by mixing the top-3 hyperparameters in~\cref{tab:pretrain}, we report the results of training models on each individual S-AF and AF dataset in~\cref{tab:single_or_mix}. The models are trained for 0.2M iterations. We note that the top hyperparameters sets are selected according to the search, where the model is trained for 40K iterations, and here we report the results of model trained for 0.2M iterations, so the top-1 hyperparameters might not have the lowest AEPE for AF models.

Unlike supervised AutoFlow, the results of S-AF trained on the top-2 hyperparameters on Sintel Final and S-AF trained on the top-3 hyperparameters on KITTI show that there is no guarantee that the top candidates returned by self-supervised AutoFlow are the optimal set of hyperparameters.
Mixing the top-3 datasets decreases the likelihood of sampling a poor-performing AutoFlow hyperparameters and improves the robustness of the algorithm.

\begin{table}[H] 
\centering
\caption{ \textbf{Training by individual S-AF dataset and mixed S-AF dataset.} We show that training on mix-3 datasets decreases the likelihood of sampling a poor-performing AutoFlow hyperparameters and improves the robustness of the algorithm.
}
\small
    \begin{tabular}{l cccc c cccc c cccc}
        & \multicolumn{4}{c}{Sintel Clean~\cite{ButlerECCV2012}} & \phantom{aaa} & \multicolumn{4}{c}{Sintel Final~\cite{ButlerECCV2012}} & \phantom{aaa} & \multicolumn{4}{c}{KITTI 2015~\cite{KITTI2015}} \\
         \cmidrule{2-5} \cmidrule{7-10} \cmidrule{12-15}     
        Method & top-1 &  top-2 & top-3 & mix-3 && top-1 &  top-2 & top-3 & mix-3 && top-1 &  top-2 & top-3 & mix-3 \\
        \midrule
\autoflow-mix~(0.2M)  & 2.11 & 2.18 & 2.10 & 2.18 && 2.85 & 2.83 & 2.82 & 2.83 && 4.70 & 4.35 & 4.58 & 4.43\\
\selfautoflow~(0.2M) & 2.16 & 2.14 & 2.13 & 2.22 && 2.83 & \bf 2.93 & 2.84 & 2.84 && 4.65 & 4.06 & \bf 5.40 & 4.58 \\
    \end{tabular}
    
\label{tab:single_or_mix}
\end{table}

\subsection{Sequence losses in the search metric of S-AF}    
    
\begin{figure}[H] 
  \centering
  \includegraphics[width=1.0\linewidth]{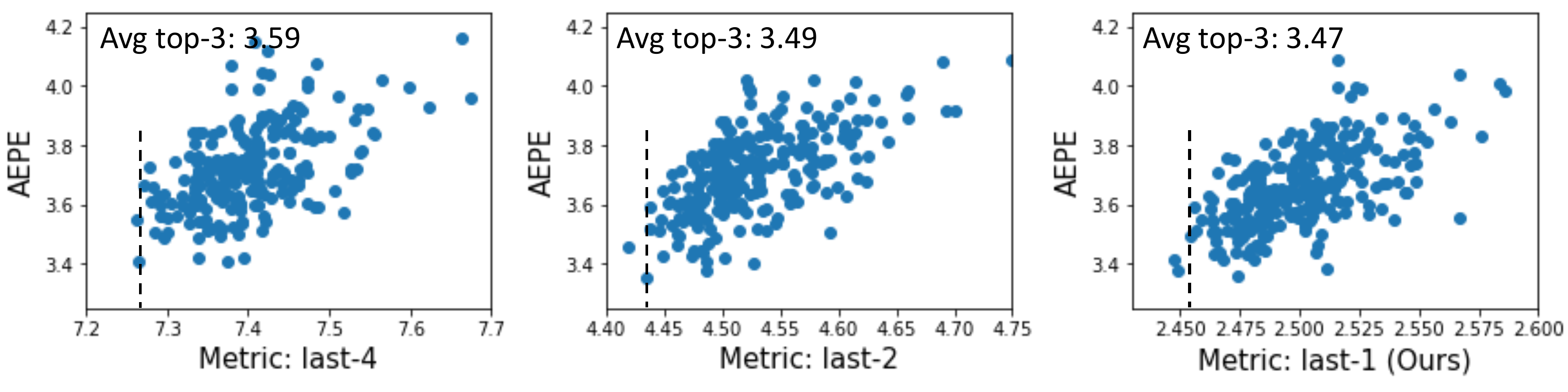}
\caption{\textbf{Sequence losses.} We find that using the intermediate predictions of RAFT to compute the search metric does not lead to a better set of S-AF hyperparameters.}
\label{fig:sequence_loss}
\end{figure}

In~\cref{sec:autoflow}, we mention that since there is no backpropagation to the model
in the search of AutoFlow, the search metric uses only the final flow prediction of RAFT instead of all intermediate. 
In~\cref{fig:sequence_loss}, we conduct a study of using the intermediate predictions of RAFT to compute the search metric.
Specifically, we compute the search metric once for each intermediate prediction and we exponentially decay the weight for earlier predictions~\cite{stone2021smurf}.
Since the search metric is computed at the original resolution of the target data, we use at most the last four predictions due to memory constraints.

We conduct the S-AF search using last-1 prediction (ours), last-2 prediction and last-4 prediction as the search metric. 
We report the average AEPE of the top-3 models selected by the search metric. The models are trained for 40k iterations in the search.
Empirically, we find that using the intermediate predictions in the search metric results in a higher AEPE and does not improve the S-AF search.

\section{Analysis and Discussion}
\subsection{Motion statistics of S-AF and AF}
\begin{figure}[H]
\centering
\begin{subfigure}{.5\textwidth}
  \centering
  \includegraphics[width=\linewidth]{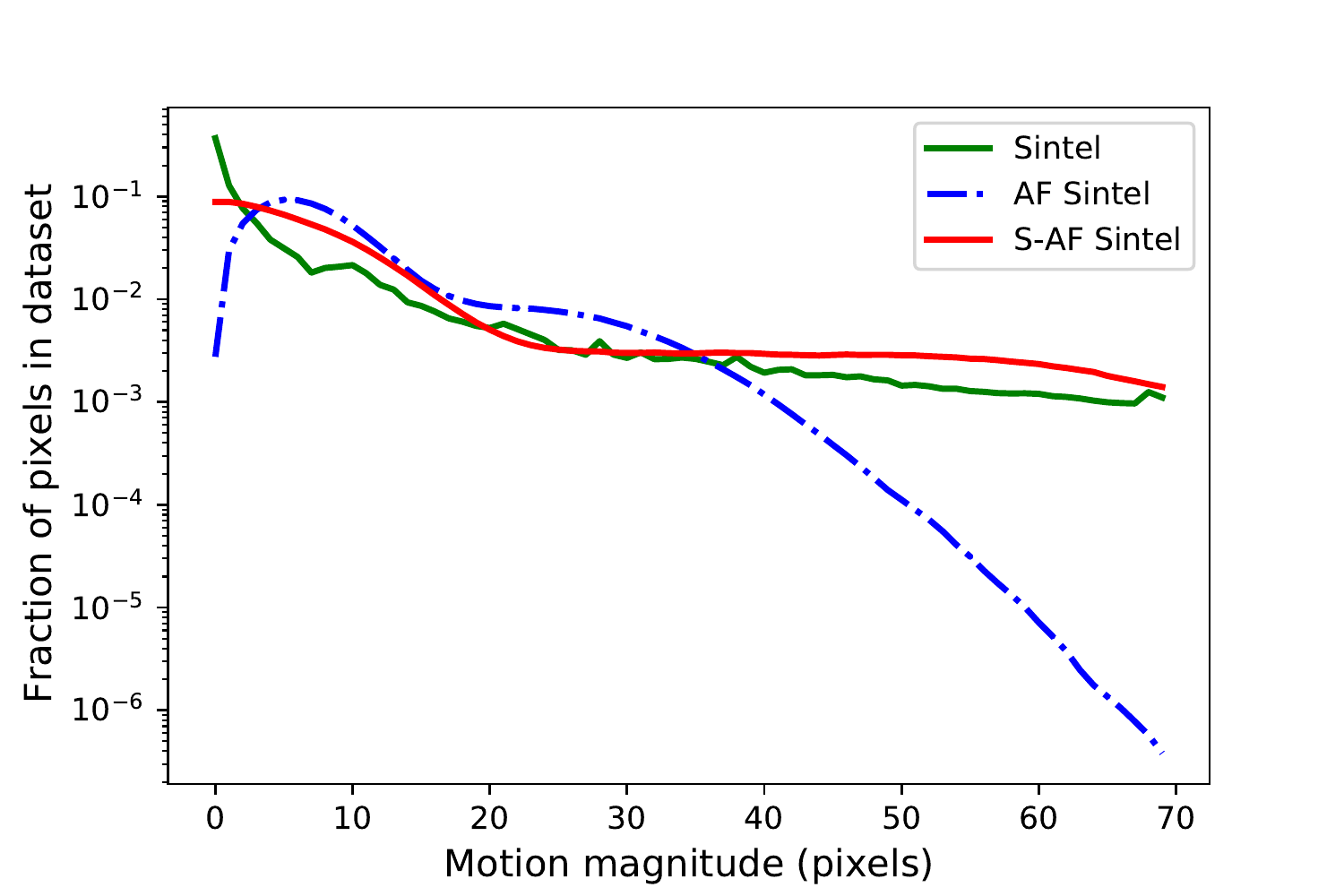}
\end{subfigure}%
\begin{subfigure}{.5\textwidth}
  \centering
  \includegraphics[width=\linewidth]{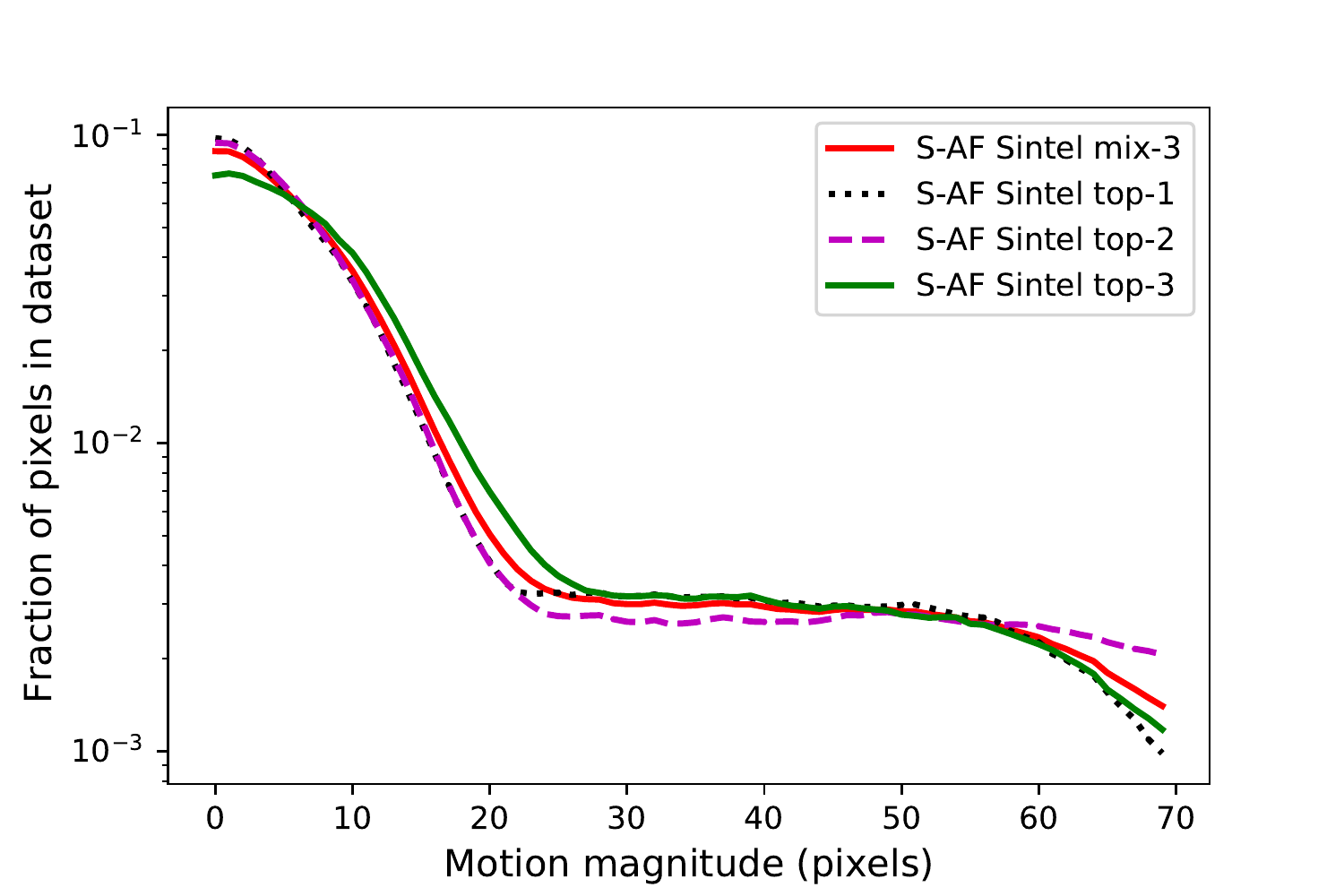}
\end{subfigure}
\caption{\textbf{Histogram of motion magnitude.} We include the motion statistics of the generated flow field by Self-AutoFlow and AutoFlow. Interestingly, the Self-AutoFlow data focuses more on small motion compared to AutoFlow. Also, the statistics of Self-AutoFlow are closer to the statistics of Sintel data. In addition, we show the statistics of individual S-AF dataset and their mixed results.}
\label{fig:motion_stats}
\end{figure}

We compute the statistics of the motion magnitude of the generated optical flow ground truth in S-AF and AF datasets in~\cref{fig:motion_stats}.
We find that when the target dataset is Sintel, the motion statistics of S-AF are similar to the statistics of Sintel data. In contrast, the motion statistics of AutoFlow are different from the Sintel data.
In addition, S-AF focuses more on the small motion compared to AF which focuses on middle-range motion. 
We hypothesize that the self-supervised search metric may have much smaller values for middle/high-range motions compared to AEPE which penalizes significantly on the error at the regions of large motions. Therefore, the S-AF data does not focus on regions with large motions compared to AF.
Similar to~\cref{tab:single_or_mix}, we also show the statistics of each individual S-AF dataset and the mixed dataset. We find the statistics are similar for each individual S-AF data.

\subsection{AEPE versus self-supervised losses of SMURF and S-AF}  

We calculate the self-supervised losses and the AEPE on the target datasets for SMURF and S-AF models in~\cref{tab:aepe_and_loss}. The losses and errors are computed for the full target datasets and we report the average. In most cases, the SMURF models have a lower self-supervised losses compared to the S-AF models, while the S-AF models have lower AEPE.

Although the self-supervised metric is highly correlated with the AEPE, optimizing it \textit{directly} by backpropagation to the model might lead to a model with lower self-supervised loss and higher EPE. In contrast, our Self-AutoFlow method uses the self-supervised loss \textit{indirectly} to assess the quality of a generated dataset, which results in a model with higher self-supervised loss and lower EPE. To conclude, Self-AutoFlow is a good strategy for using self-supervised losses.

\begin{table}[H] 
    \centering
    \caption{ {\bf Self-supervised losses versus AEPE.} We compute the photometric, distillation and smoothness loss averaged on the training set. 
    We show that our S-AF model which uses the self-supervised loss \textit{indirectly} to assess the quality of a generated dataset results in a model with higher self-supervised loss and lower EPE.
    }   
    \resizebox{0.95\textwidth}{!}{
	
    \begin{tabular}{l ccccc c ccccc}
        & \multicolumn{5}{c}{Sintel Final~\cite{ButlerECCV2012}} & \phantom{aaa} & \multicolumn{5}{c}{KITTI 2015~\cite{KITTI2015}} \\
         \cmidrule{2-6} \cmidrule{8-12}     
        Method& $\mathcal{L}_\text{photo}$ $\downarrow$ & $\mathcal{L}_\text{distill}$ $\downarrow$ & $\mathcal{L}_\text{smooth}$ $\downarrow$ & $\mathcal{L}_\text{total}$ $\downarrow$ & \textit{AEPE}$\downarrow$ & & $\mathcal{L}_\text{photo}$ $\downarrow$ & $\mathcal{L}_\text{distill}$ $\downarrow$ & $\mathcal{L}_\text{smooth}$ $\downarrow$ & $\mathcal{L}_\text{total}$ $\downarrow$ & \textit{AEPE} $\downarrow$ \\
        \midrule
        SMURF Chairs \cite{stone2021smurf} & 2.20 & 0.70 & \bf 0.013 & 2.92 & 3.35 && 2.61 & \bf 1.05 & \bf 0.0046 &  \bf 3.67 & 7.94\\ 
        \selfautoflow & 2.20 & \bf 0.44 & 0.017 &  \bf 2.66 & \bf 2.57 && \bf 2.54 & 1.24 & 0.0052 & 3.78 & \bf 4.28 \\ 
        \midrule
        +SS Sintel/KITTI \\
        SMURF \cite{stone2021smurf} & \bf 2.17 & 0.67 & \bf 0.012 &  \bf 2.86 & 2.80 && 2.54 & \bf 0.80 & 0.0046 &  \bf 3.35 & 2.01\\ 
        \selfautoflow & 2.20 & \bf 0.65 & 0.013 & 2.87 & \bf 2.40 && \bf 2.49 & 0.87 & \bf 0.0043 & 3.37 & \bf 1.94 \\ \hline
    \end{tabular}
	}
    \label{tab:aepe_and_loss}
\end{table}

\section{Additional Results}

\subsection{Visualization of keypoint propagation on BADJA}

\begin{figure}[H] 
\centering
\begin{subfigure}{1.0\textwidth}
  \centering
  \includegraphics[width=\linewidth]{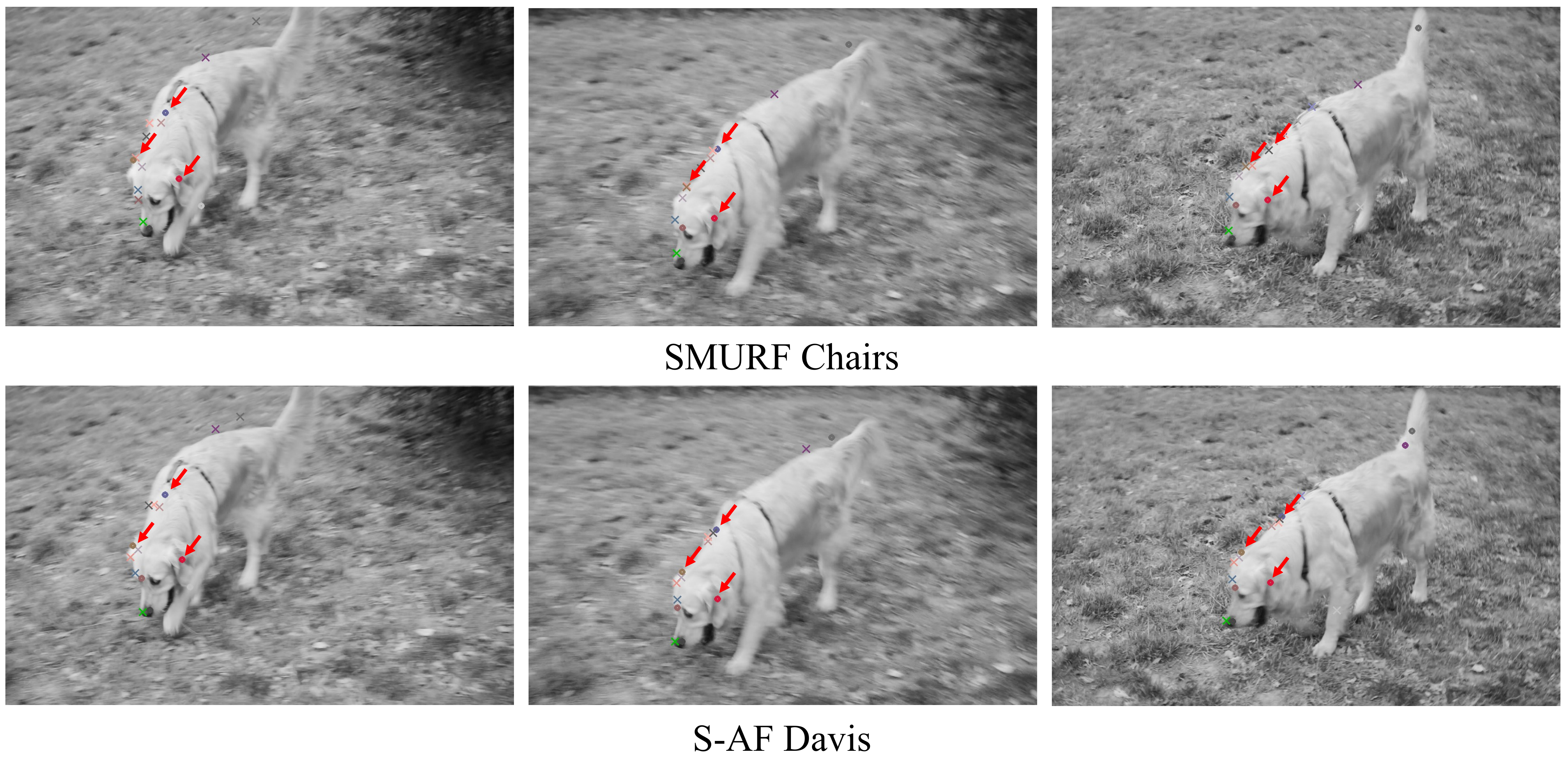}
\end{subfigure}%

\begin{subfigure}{1.0\textwidth}
  \centering
  \includegraphics[width=\linewidth]{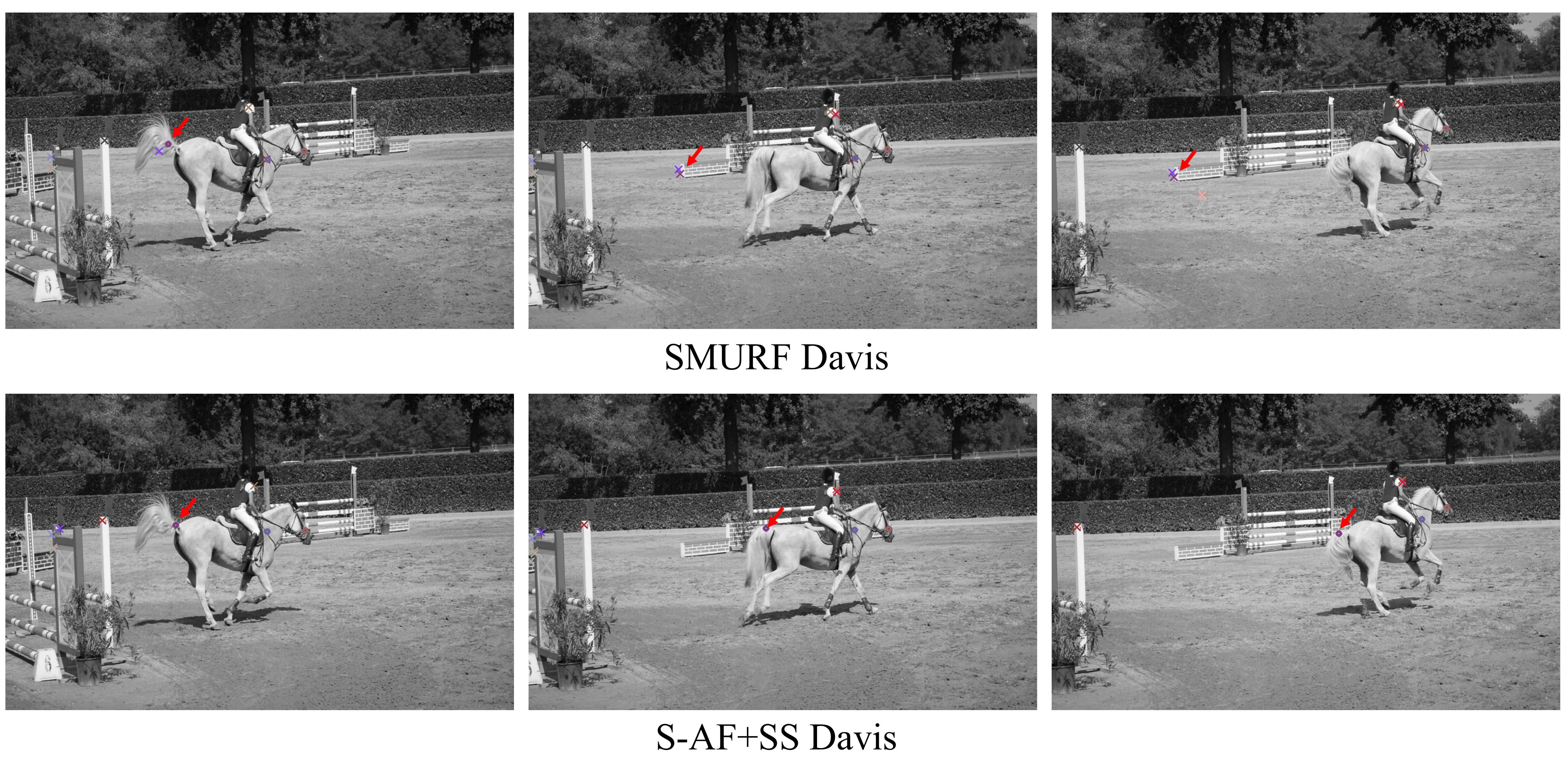}
\end{subfigure}

\caption{\textbf{Visual results of keypoints on BADJA.} 
The keypoints correctly tracked are marked as a dot, and the keypoints with the wrong trajectory are marked as a cross. Compared the results without self-supervised fine-tuning in (a), S-AF track the three keypoints (gray, red, and brown) correctly, while SMURF loses the brown keypoint in the second frame and the gray keypoint in the third frame. As for the results with self-supervised fine-tuning, S-AF correctly tracks the purple keypoint on the tail, and SMURF loses it since the second frame.
}
\label{fig:badja_vis}
\end{figure}

We visualize the keypoint propagation results on BADJA sequences by SMURF and our S-AF in~\cref{fig:badja_vis}.
The keypoints correctly propagated are marked as a dot, and the keypoints with the wrong predicted trajectory are marked as a cross. 
Compared the results without self-supervised fine-tuning, S-AF tracks the three keypoints on the back (gray), left ear (red), and right ear (brown) correctly. On the other hand, SMURF loses the keypoint on the right ear (brown) since the second frame and loses the keypoint on the back (gray) since the third frame in the dog sequence. 
As for the models with self-supervised fine-tuning, we show the keypoint in the horsejump-low sequence. S-AF correctly predicts the trajectory of the purple keypoint on the tail, while SMURF loses it since the second frame.

\subsection{Benchmark results}
\begin{figure}[H]
\centering
\begin{subfigure}{.8\textwidth}
  \centering
  \includegraphics[width=\linewidth]{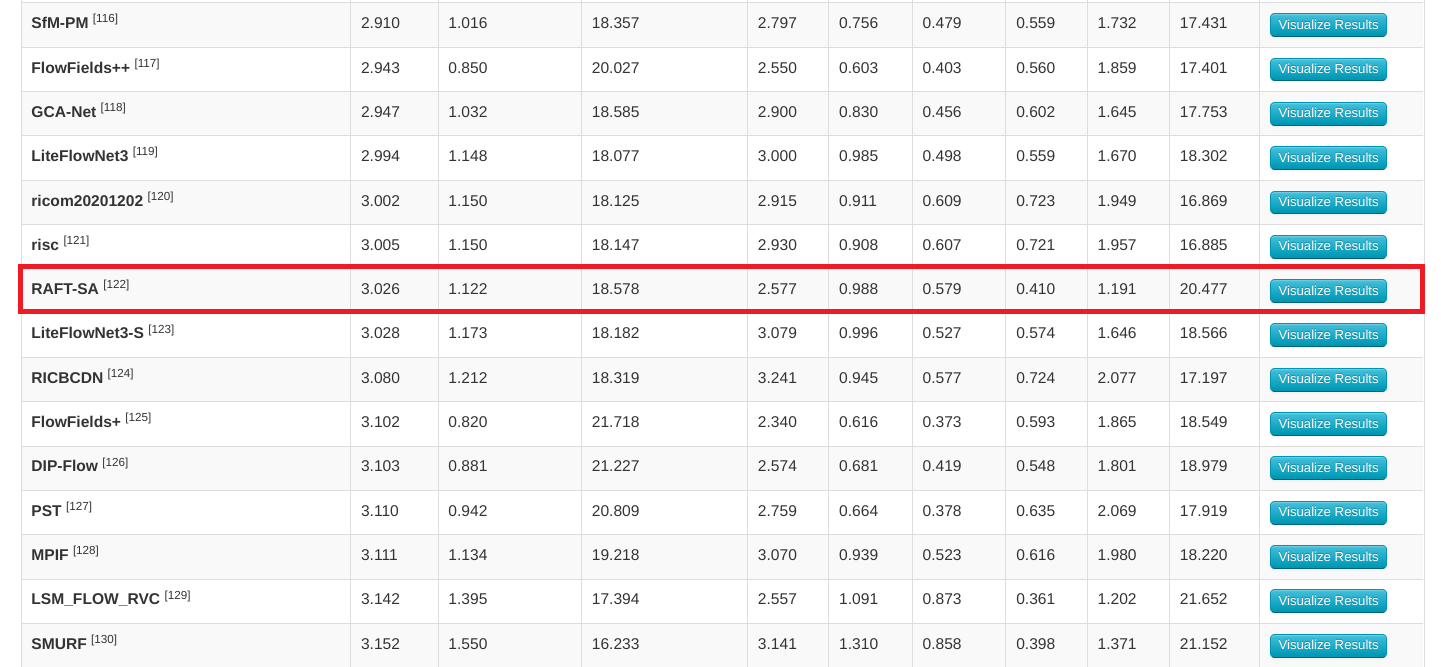}
  \caption{Sintel Clean}
\end{subfigure}%

\begin{subfigure}{.8\textwidth}
  \centering
  \includegraphics[width=\linewidth]{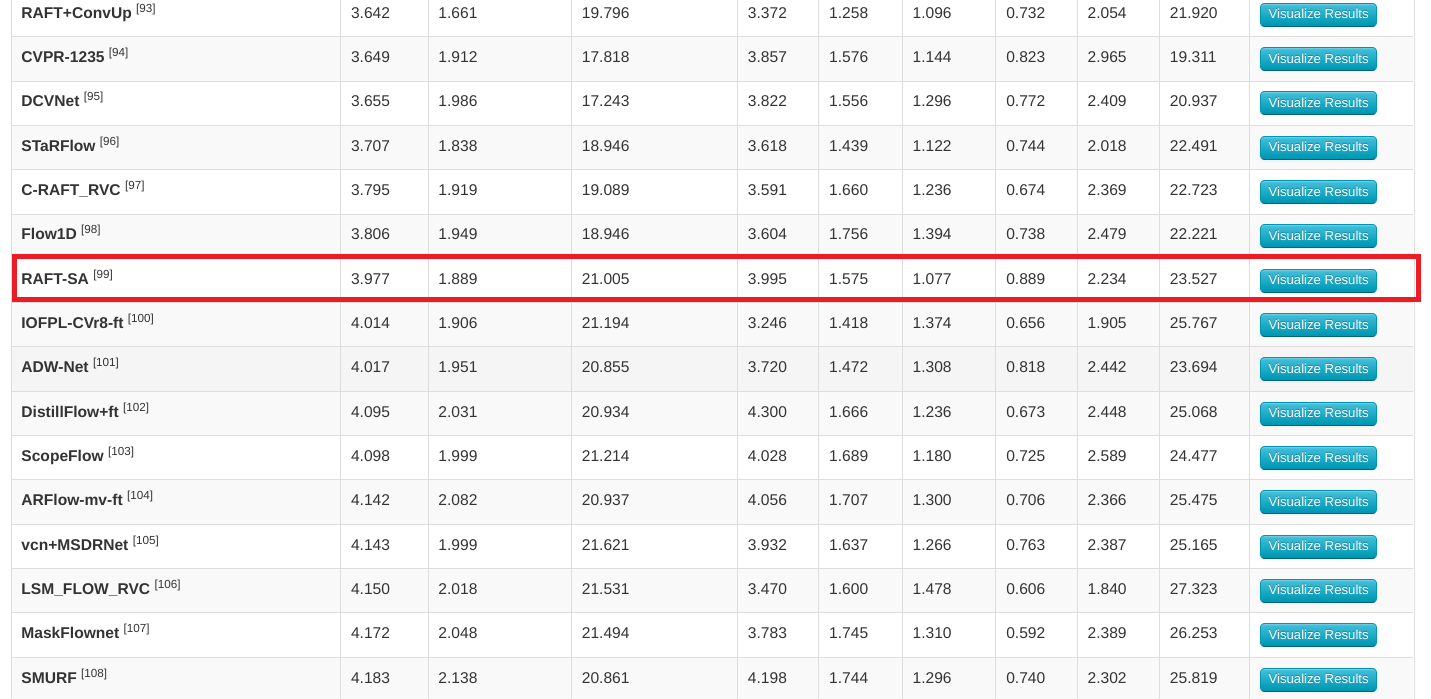}
  \caption{Sintel Final}
\end{subfigure}

\begin{subfigure}{.8\textwidth}
  \centering
  \includegraphics[width=\linewidth]{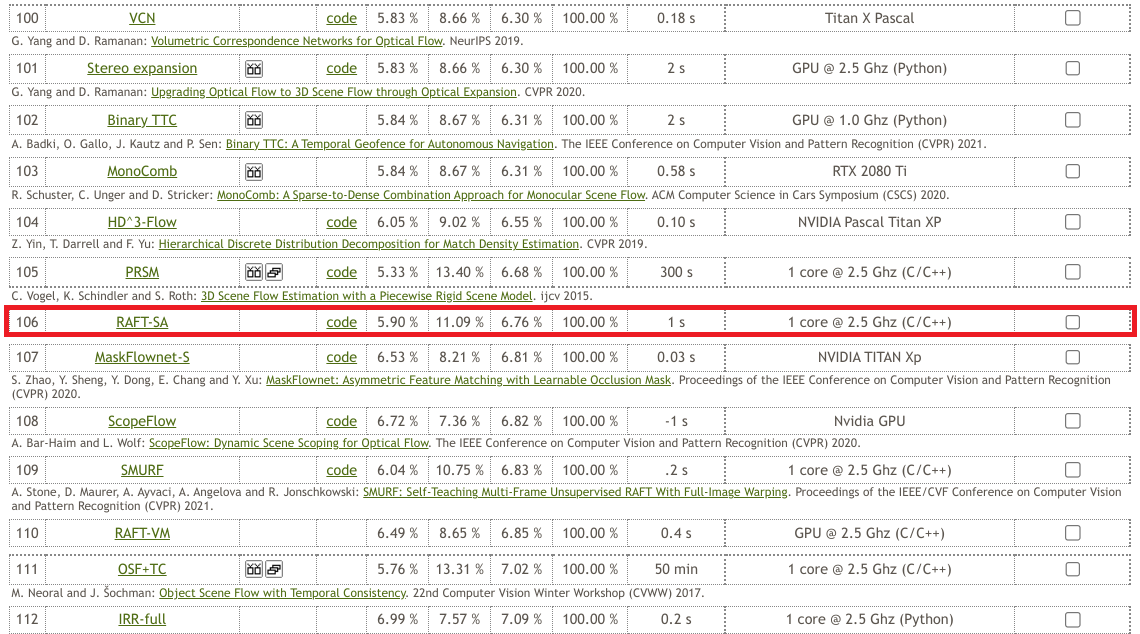}
  \caption{KITTI}
\end{subfigure}
\caption{\textbf{Screenshot of S-AF+SS on public benchmark.} Our method was temporarily named as RAFT-SA.}
\label{fig:screen_sota}
\end{figure}

\begin{figure}[H]
\centering
\begin{subfigure}{.8\textwidth}
  \centering
  \includegraphics[width=\linewidth]{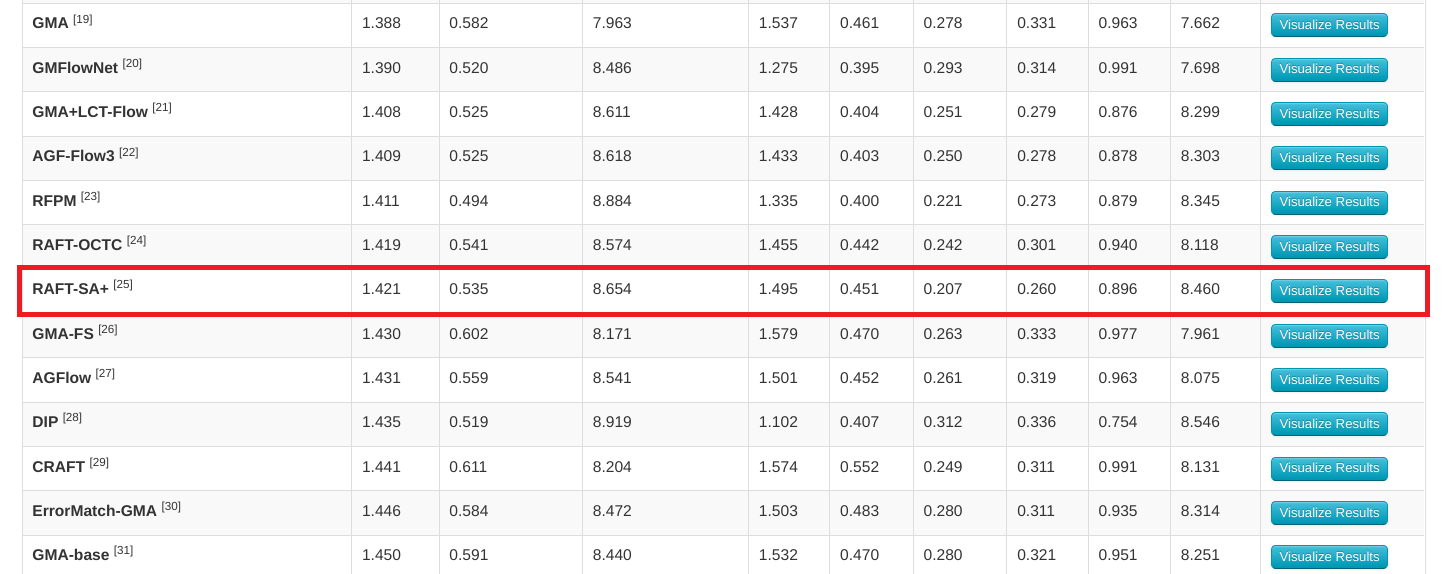}
  \caption{Sintel Clean}
\end{subfigure}%

\begin{subfigure}{.8\textwidth}
  \centering
  \includegraphics[width=\linewidth]{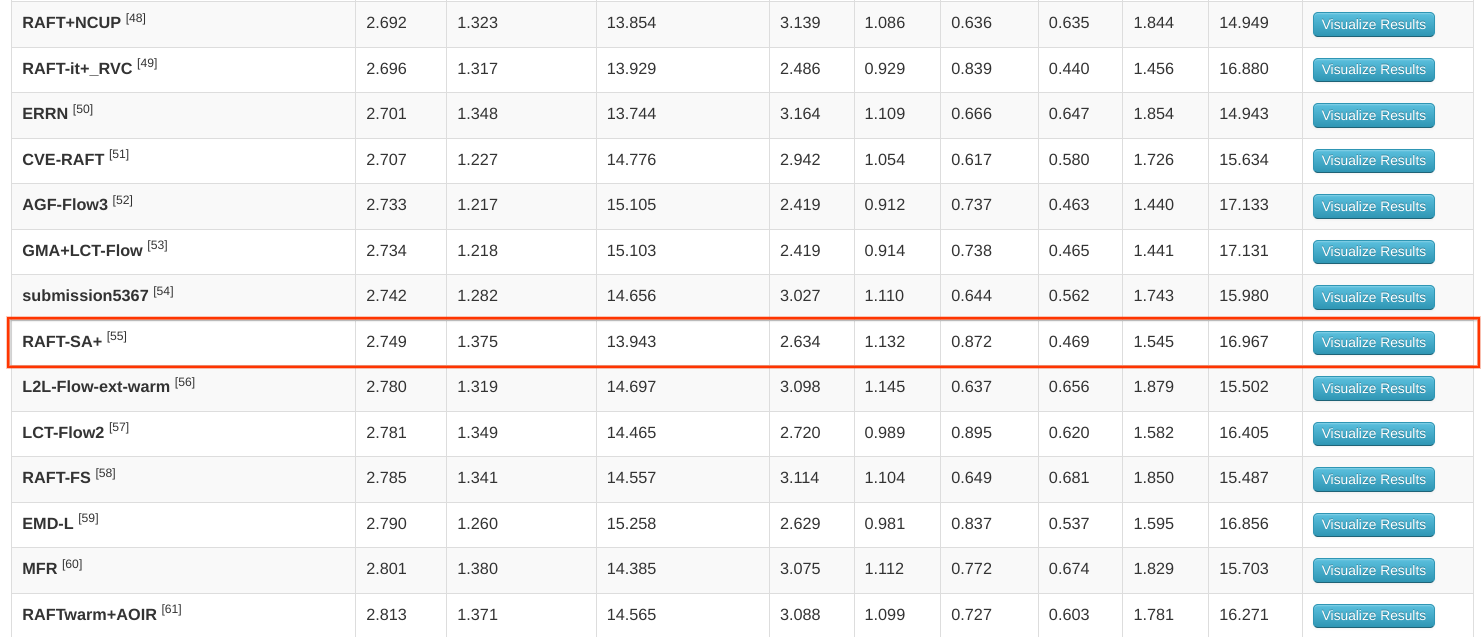}
  \caption{Sintel Final}
\end{subfigure}

\begin{subfigure}{.8\textwidth}
  \centering
  \includegraphics[width=\linewidth]{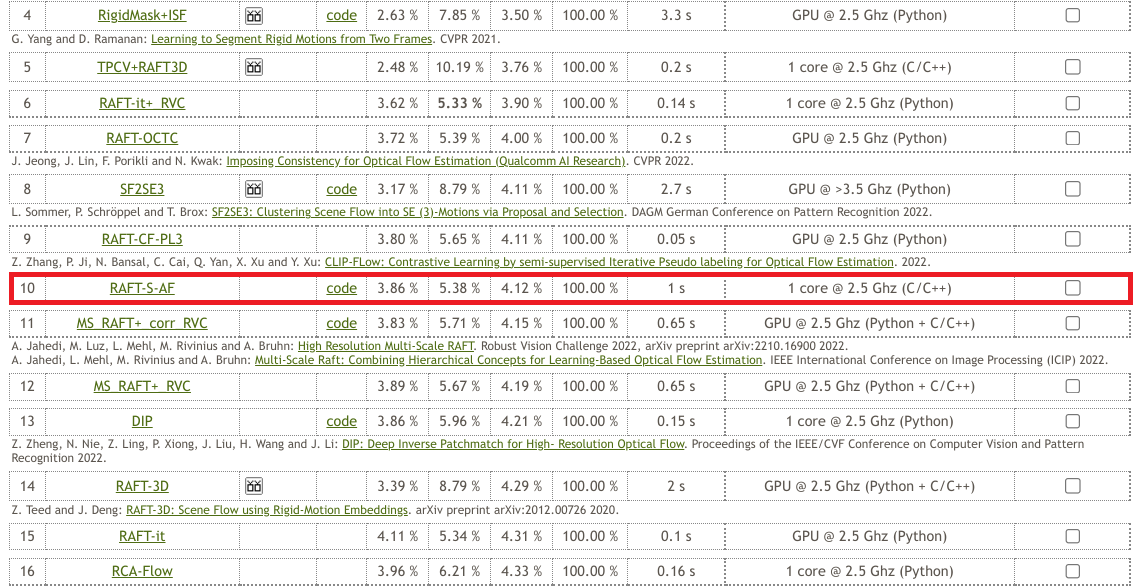}
  \caption{KITTI}
\end{subfigure}
\caption{\textbf{Screenshot of the supervised fine-tuning results of S-AF on public benchmark.} Our method was temporarily named as RAFT-SA+ and RAFT-S-AF.}
\label{fig:screen_benchmark}
\end{figure}

We provide the screenshots of both models on the public benchmarks in~\cref{fig:screen_sota} and~\cref{fig:screen_benchmark}.
As listed in~\cref{tab:sota}, we provide the detailed performance of our S-AF+SS models on public benchmarks in~\cref{tab:sota_details}. 
The S-AF+SS model is more accurate in most cases while less accurate on unmatch and s0-10 for Sintel benchmark, and on F1-fg all for KITTI benchmark compared to SMURF.
As shown in~\cref{tab:benchmark}, we show the detailed performance of the supervised fine-tuning model RAFT-S-AF in~\cref{tab:ft_details}.
For KITTI benchmark, RAFT-S-AF is more accurate in most cases while less accurate for F1-fg. RAFT-S-AF is more accurate for all cases for Sintel Clean and Sintel Final.

\begin{table}[H]
\begin{center}
\small
\caption{{Detailed performance of S-AF+SS on public benchmark.} }
\begin{tabular}{ l|ccccccccc } 
{\bf Model} & {\bf all}    & {\bf match} & {\bf unmatch}    & {\bf d0-10} &    {\bf d10-60} &    {\bf d60-140} &    {\bf s0-10} & {\bf s10-40} &    {\bf s40+} \\ \hline
SMURF & 3.15	& 1.55	& 16.23	& 3.14	& 1.31	& 0.86	& 0.40	& 1.37	& 21.15 \\
S-AF+SS & 3.03	& 1.12	& 18.58	& 2.58	& 0.99	& 0.58	& 0.41	& 1.19	& 20.48 \\
\end{tabular}
\vspace{1mm}
\\ (a) Sintel Clean \\
\vspace{1mm}
\begin{tabular}{ l|ccccccccc } 
{\bf Model} & {\bf all}    & {\bf match} & {\bf unmatch}    & {\bf d0-10} &    {\bf d10-60} &    {\bf d60-140} &    {\bf s0-10} & {\bf s10-40} &    {\bf s40+} \\ \hline
SMURF & 4.18	& 2.14	& 20.86	& 4.20	& 1.74	& 1.30	& 0.74	& 2.30	& 25.82 \\
S-AF+SS & 3.98	& 1.89	& 21.01	& 4.00	& 1.58	& 1.08	& 0.89	& 2.23	& 23.53 \\
\end{tabular}
\vspace{1mm}
\\ (b) Sintel Final \\
\vspace{1mm}
\begin{tabular}{l|c c  c c c c c}
\multirow{2}{*}{{\bf Model}} & \multicolumn{3}{c}{All} & \multicolumn{3}{c}{Occ} \\
 & {\bf Fl-bg} & {\bf Fl-fg} & {\bf Fl-all} & {\bf Fl-bg} & {\bf Fl-fg} & {\bf Fl-all} \\ \hline
SMURF & 6.04 \%	& 10.75 \%	& 6.83 \% & 4.46 \%	& 8.86 \%	& 5.26 \% \\
S-AF+SS & 5.90 \%	& 11.09 \%	& 6.76 \% & 4.41 \%	& 8.67 \%	& 5.18 \% \\
\end{tabular}
\vspace{1mm}
\\ (c) KITTI \\
\label{tab:sota_details}
\end{center}
\end{table}

\begin{table}[H]
\begin{center}
\small
\caption{{Detailed performance of the supervised fine-tuning results of S-AF on public benchmark.}}
\begin{tabular}{ l|ccccccccc } 
{\bf Model} & {\bf all}    & {\bf match} & {\bf unmatch}    & {\bf d0-10} &    {\bf d10-60} &    {\bf d60-140} &    {\bf s0-10} & {\bf s10-40} &    {\bf s40+} \\ \hline
RAFT-it &1.55 &	0.61 &	9.24 &	1.66 &	0.51 &	0.27 &	0.29 &	0.97 &	9.26 \\
RAFT-S-AF &1.42 &	0.54 &	8.65 &	1.50 &	0.45 &	0.21 &	0.26 &	0.90 &	8.46 \\ 
\end{tabular}
\vspace{1mm}
\\ (a) Sintel Clean \\
\vspace{1mm}
\begin{tabular}{ l|ccccccccc } 
{\bf Model} & {\bf all}    & {\bf match} & {\bf unmatch}    & {\bf d0-10} &    {\bf d10-60} &    {\bf d60-140} &    {\bf s0-10} & {\bf s10-40} &    {\bf s40+} \\ \hline
RAFT-it &2.90    &1.41    &15.03    &2.81    &1.16    &0.88    &0.51    &1.70    &17.62 \\
RAFT-S-AF & 2.75 & 1.38 & 13.94 & 2.63 & 1.13 & 0.87 & 0.47 & 1.55 & 16.97  \\
\end{tabular}
\vspace{1mm}
\\ (b) Sintel Final \\
\vspace{1mm}
\begin{tabular}{l|c c  c c c c c}
\multirow{2}{*}{{\bf Model}} & \multicolumn{3}{c}{All} & \multicolumn{3}{c}{Occ} \\
 & {\bf Fl-bg} & {\bf Fl-fg} & {\bf Fl-all} & {\bf Fl-bg} & {\bf Fl-fg} & {\bf Fl-all} \\ \hline
RAFT-it &  4.11 \% & 5.34 \% & 4.31 \% & 2.68 \% & 2.77 \% & 2.70 \%\\ 
RAFT-S-AF &  3.86 \% & 5.38 \% & 4.12 \% & 2.52 \% & 2.86 \% & 2.59 \%\\ 
\end{tabular}
\vspace{1mm}
\\ (c) KITTI \\

\label{tab:ft_details}
\end{center}
\end{table}

\end{document}